\documentclass[11pt]{article}
\usepackage{latexsym}
\usepackage{theorem}
\usepackage{graphicx}
\usepackage{amsmath}
\usepackage{color}
\usepackage{xcolor}
\usepackage{amsfonts}
\usepackage{enumitem}
\usepackage[page]{appendix}
\usepackage{natbib}
\usepackage{pdf14}
\usepackage{soul}
\usepackage{hyperref} 
\usepackage{mathtools}
\DeclarePairedDelimiter{\floor}{\lfloor}{\rfloor}

\usepackage[margin=1in]{geometry}

\theorembodyfont{\rmfamily}
\newtheorem{theorem}{Theorem}
\newtheorem{conjecture}[theorem]{Conjecture}
\newtheorem{lemma}[theorem]{Lemma}
\newtheorem{proposition}[theorem]{Proposition}

\newtheorem{definition}[theorem]{Definition}
\newenvironment{proof}[1][Proof]{\noindent\textbf{#1.} }{\ \rule{0.5em}{0.5em}}

\DeclareMathOperator*{\argmax}{arg\,max}

\theoremstyle{break}
\newtheorem{algorithm}[theorem]{Algorithm}

\newtheorem{example}[theorem]{Example}

\newcommand{\be}{\beta}

\newcommand{\red}[1]{{\color{red}#1}}

\graphicspath{{../fig/}}

\title{A Classical Search Game in Discrete Locations}

\author{Jake Clarkson\thanks{STOR-i Centre for Doctoral Training, Science and Technology Building, Lancaster University, LA1 4YR, j.clarkson@lancaster.ac.uk} \and Kyle Y. Lin\thanks{Operations Research Department, Naval Postgraduate School, Monterey, CA 93943, kylin@nps.edu} \and Kevin D. Glazebrook\thanks{Department of Management Science, Lancaster University, LA1 4YX, k.glazebrook@lancaster.ac.uk}}

\begin{document}

\maketitle

\begin{abstract}
\noindent
Consider a two-person zero-sum search game between a hider and a searcher.
The hider hides among $n$ discrete locations, and the searcher successively visits individual locations until finding the hider. 
Known to both players, a search at location $i$ takes $t_i$ time units and detects the hider---if hidden there---independently with probability $q_i$, for $i=1,\ldots,n$.
The hider aims to maximize the expected time until detection, while the searcher aims to minimize it.
We prove the existence of an optimal strategy for each player.
In particular, the hider's optimal mixed strategy hides in each location with a nonzero probability, and the searcher's optimal mixed strategy can be constructed with up to $n$ simple search sequences.
We develop an algorithm to compute an optimal strategy for each player, and compare the optimal hiding strategy with the simple hiding strategy which gives the searcher no location preference at the outset.
\end{abstract}


\section{Introduction}
Consider the following two-person zero-sum game. A hider chooses one of $n$ locations (henceforth boxes for conciseness) to hide in, and a searcher searches these boxes one at a time in order to find the hider.
A search in box $i$ takes $t_i > 0$ time units and will find the hider with probability $q_i \in (0, 1)$ if the hider is there, for $i=1,\ldots, n$.
Due to the possibility of overlook, the searcher may need to visit a box many times to find the hider, and the total time until detection can be arbitrarily long.
The searcher wants to minimize the expected total time until the hider is found, while the hider wants to maximize it.

If the hider announces to the searcher the probability with which they will hide in each box at the beginning of the search, then the resulting search model is one that is well studied in the literature.
An optimal search strategy, first discovered by Blackwell (reported in \cite{Matula}), is to always next search a box with a maximal probability of detection per unit time at that moment.
In other words, if presently the hider is believed to be in box $i$ with probability $p'_i$, $i=1,\ldots,n$---a value updated throughout the search using Bayes' theorem---then it is optimal to next search a box with a maximal $p'_i q_i /t_i$.
A comment by Kelly in \cite{Gittins79} notes that Blackwell's solution is equivalent to a Gittins index policy obtained by modelling the search as a tractable version of the multi-armed bandit problem~\citep{KGbook}.
Other variations of this search model have been studied in \cite{Ross:1969cm, Kadane:1971wj, ChewJr:1973td, Wegener:1980hv, Kress:2008co}.

The search problem becomes a substantially more complicated \emph{search game} if the 
searcher does not know the hiding strategy.
Whilst the hider has $n$ pure strategies to choose from---each corresponding to hiding in a box---a pure search strategy must specify an indefinite, ordered sequence of boxes for the searcher to search, because the search can take arbitrarily long.

The special case of our search game with $t_i=1$ for $i=1,\ldots,n$---the case of unit search time---has been studied in the literature with limited results.
\cite{Bram1963} proves an optimal search strategy exists, and \cite{Ruckle:1991} solves a few special cases and finds the best pure search strategy. \cite{GitRob1} and \cite{Gittins:1989} further specialize to $n=2$ boxes; the former finds an optimal hiding strategy under certain conditions, while the latter shows the existence of an optimal search strategy that randomly chooses between just two simple search sequences. \cite{GitRob2} develops an algorithm to estimate an optimal hiding strategy for $n \geq 3$ boxes. \cite{Subelman} studies a different objective function in which the searcher wants to maximize the probability of finding the hider by an announced deadline, while the hider wants to minimize it.
\cite{Lin2016} extends the results in \cite{Subelman} to show that the searcher has a uniformly optimal strategy that maximizes the probability of finding the hider simultaneously for all deadlines.

The main contribution of this paper is to develop a rigorous mathematical framework to extend earlier results to the search game in its full generality.
In particular, we allow for an arbitrary number of boxes $n$, each having its own search time $t_i>0$,
$i=1,\ldots,n$.
We first prove that the value of the game exists and each player has an optimal strategy.
We next develop properties of the searcher's optimal strategies, and show that the searcher can construct an optimal strategy by randomly choosing between $n$ simple search sequences.
Based on these properties, we present a practical procedure to test the optimality of a hiding strategy and an algorithm to estimate each player's optimal strategy by successively bounding the value of the game. The findings in this paper both strengthen our understanding of the search game of interest and provide insight into effective practice in real-world search for an intruder.

Our work falls in the general area of search theory, where a searcher seeks a hidden target. Besides the aforementioned papers closely related to our work, search theory has a rich literature with a variety of search models.
Common choices of search spaces include the real line, a two-dimensional area, or a network of nodes connected by edges.
The target may be stationary, or move around the search space via either a known or random path. Some works assume that the searcher detects the target when their paths cross, and some others consider the possibility of overlook. The searcher may aim to find the target as soon as possible, or maximize the probability of detection before a deadline.
For a general review of search theory, see \cite{Washburn:2002tw}, \cite{Stone2004} and \cite{Stone2016}.
For a summary of search games in which the target is a hider actively trying to avoid detection in a time-stationary search space, see Book 1 of \cite{AlpernGal} and Part I of \cite{Alpern4auth}. See \cite{Garrec:2020} for a novel stochastic search game where the search space, a network, changes randomly over time.
If the hider wishes to be found, such as a survivor of a disaster, see rendezvous search in Book 2 of \cite{AlpernGal}.

The rest of the paper proceeds as follows.
Section~\ref{sec:modelprelim} formulates our search game as a semi-finite two-person zero-sum game and establishes the existence of the value of the game and an optimal strategy for the hider.
Section~\ref{sec:OSSexist} proves the existence of an optimal strategy for the searcher, and Section \ref{sec:properties} shows that there exists an optimal search strategy which involves randomly choosing among $n$ simple search sequences.
Section~\ref{sec:findopt} develops a practical test to determine whether a given hiding strategy is optimal by solving a finite two-person zero-sum game, and presents an algorithm to estimate each player's optimal strategy by successively calculating tighter bounds on the value of the game.
Section~\ref{sec:numerical} presents numerical results.
Section~\ref{sec:conclusion} concludes and offers some future research directions.

\section{Model and Preliminaries} \label{sec:modelprelim}
Consider a two-person zero-sum \emph{search game} $G$ as follows.
A hider decides where to hide among $n$ boxes labelled $1,\ldots , n$, and a searcher decides an ordered sequence of boxes to search.
A search in box $i$ takes time $t_i$ and will find the hider with detection probability $q_i$, $i=1,\ldots,n$, if the hider is indeed hidden there. These quantities are common knowledge to both players.
The game proceeds until the searcher finds the hider, with the total search time the payoff of the game.
The searcher wishes to minimize the expected payoff---namely the expected time to detection---while the hider wishes to maximize it.

The hider's pure strategy space is $\{1, \ldots, n\}$, where each pure strategy corresponds to a box in which to hide.
A mixed hiding strategy is a probability vector $\mathbf{p} \equiv (p_1,\ldots, p_n)$ such that the hider hides in box $i$ with probability $p_i$, where $0 \leq p_i \leq 1$ for $i = 1,\ldots, n$, and $\sum_{i=1}^n p_i = 1$.
The searcher's pure strategy space is the infinite Cartesian product $\mathcal{C} \equiv \{1,2,\ldots,n\}^\infty$.
Each pure strategy is a \emph{search sequence}---an infinite, ordered list of boxes to search until the hider is found.
A mixed search strategy is a function $\eta$ with domain $\mathcal{C}$ such that the set $\{\xi \in \mathcal{C} : \eta(\xi)>0\}$ is countable, and $\sum_{\xi \in \mathcal{C}} \eta(\xi) = 1$.
Under strategy $\eta$, the searcher plays search sequence $\xi \in \mathcal{C}$ with probability $\eta(\xi)$, and we say $\eta$ is a \emph{mixture} of those $\xi$ with $\eta(\xi)>0$.

For a search sequence $\xi \in \mathcal{C}$, write $V_i(\xi)$ for the expected time to detection if the hider hides in box $i$, for $i = 1,\ldots, n$.
In other words, $V_i(\xi)$ is the expected payoff for the hider-searcher strategy pair $(i, \xi)$.
While the hider's pure strategy space is of size $n$, the searcher's pure strategy space $\mathcal{C}$ is uncountable; 
therefore, $G$ is a two-person zero-sum semi-finite game.

The hider seeks a mixed strategy to guarantee the highest possible expected time to detection regardless of what the searcher does, so seeks to determine
\begin{equation}
\text{(Hider)} \qquad
v_1 \equiv \max_{\mathbf{p}} \inf_{\xi \in \mathcal{S}} \sum_{i=1}^n p_i V_i(\xi).
\label{eq:hider}
\end{equation}
Likewise, the searcher seeks to determine
\begin{equation*}
\text{(Searcher)} \qquad
v_2 \equiv \inf_{\eta} \max_{i \in \{1,\ldots, n\}} \sum_{\xi \in \mathcal{C}} V_i(\xi) \eta(\xi).
\end{equation*}
By definition, it is clear that $v_1\leq v_2$. Using the standard results for semi-finite games (see, for example, Chapter 13 in \cite{Ferguson2020}), we can establish the following.
Because the payoff function---namely the time to detection---is bounded below by 0, it follows that $v_1=v_2$, which is the value of $G$, written by $v^*$. In addition, the hider has an optimal strategy that guarantees an expected time to detection of at least $v^*$, and the searcher has an $\epsilon$-optimal strategy; that is, for an arbitrarily small $\epsilon > 0$, the searcher can find a strategy to guarantee an expected time to detection of at most $v^*+\epsilon$. The next section is dedicated to showing that the searcher does have an optimal strategy.

\section{Existence of Optimal Search Strategies} \label{sec:OSSexist}
The aim of this section is to prove that the searcher has an optimal strategy guaranteeing an expected time to detection of at most $v^*$ in the search game $G$.
We begin by reformulating $G$ as an $\mathcal{S}$-game of \cite{Blackwell1954}, in which, instead of choosing a pure strategy in $\mathcal{C}$, the searcher chooses a vector in the set
\begin{equation*}
\mathcal{S} \equiv \{(V_1(\xi),\ldots, V_n(\xi)): \xi \in \mathcal{C}\} \subset \mathbb{R}^n.
\end{equation*}
If the hider hides in box $i \in \{1, \ldots , n\}$ and the searcher selects $(V_1(\xi),\ldots, V_n(\xi)) \in \mathcal{S}$, then the expected payoff is $V_i(\xi)$.

By Theorem 2.4.1 of \cite{Blackwell1954}, the searcher selecting a mixed strategy is equivalent to choosing a point in $\text{Conv}(\mathcal{S})$, the convex hull of $\mathcal{S}$.
By Theorem 2.4.2, if $\mathcal{S}$, or equivalently $\text{Conv}(\mathcal{S})$, is closed, then there exists an optimal search strategy which is a mixture of at most $n$ search sequences.


The intuition behind this result is the following, adapted from Chapter 13 of \cite{Ferguson2020}. If $\mathbf{s} \equiv (s_1, \ldots, s_n) \in \text{Conv}(\mathcal{S})$, then there exists a mixed search strategy which, if the hider hides in box $i$, achieves an expected payoff $s_i$, $i =1,\ldots , n$. It follows that the value of the game $v^*$ satisfies
\begin{equation} \label{eqn:valuesgame}
v^*=\inf_{\mathbf{s} \in \text{Conv}(\mathcal{S})} \left\{ \max_{i \in \{1,\ldots , n\}} s_i  \right\}.
\end{equation}
If $\text{Conv}(\mathcal{S})$ is closed, then the infimum in \eqref{eqn:valuesgame} is attained, so there exists $\mathbf{s}^* \equiv (s_1^*, \ldots, s_n^*) \in \text{Conv}(\mathcal{S})$ with $\max_{i \in \{1,\ldots , n\}} s_i^*=v^*$; it follows that $\mathbf{s}^*$ is an optimal search strategy. See \cite{Ruckle:1991} for more on the geometrical interpretation of optimal strategies in the search game, particularly for $n=2$.

\cite{Bram1963} concluded that an optimal search strategy exists for the search game with $t_i=1$ for $i=1,\ldots , n$ by showing that $\text{Conv}(\mathcal{S})$ is closed. In this section, we take a different approach to extend the result to arbitrary $t_i > 0$, $i=1,\ldots, n$.


\subsection{Preliminary Properties of Optimal Strategies} \label{sec:propsforproof}

Recall that a pure search strategy is a search sequence---an infinite, ordered list of boxes. 
We begin by defining a particular type of search sequence.


\begin{definition} \label{def:GIP}
A \emph{Gittins search sequence} against a mixed hiding strategy $\mathbf{p} \equiv (p_1, \ldots, p_n)$ is an infinite, ordered list of boxes that meets the following rule. If $m_i \in \{1,2,\ldots\}$ searches have already been made of box $i$ during the search process for $i = 1, \ldots , n$, then the next search is some box $j$ satisfying
\begin{equation} \label{eq:simpleGI}
j = \argmax_{i \in \{1, \ldots ,  n\}} \frac{p_i(1-q_i)^{m_i}q_i}{t_i}.
\end{equation}
\end{definition}
The terms in \eqref{eq:simpleGI} are known as \emph{Gittins indices}, and a Gittins search sequence always next searches a box with a maximal Gittins index. 
There may be multiple Gittins search sequences against the same hiding strategy $\textbf{p}$ due to ties for the maximum in \eqref{eq:simpleGI}, so we write $\mathcal{C}_{\mathbf{p}} \subset \mathcal{C}$ for the set of Gittins search sequences against $\mathbf{p}$.  

If the searcher \emph{knows} that the hider will choose mixed strategy $\mathbf{p}$, 
several authors (\cite{Norris1962}, \cite{Bram1963}, Blackwell (reported in \cite{Matula}), \cite{Black:1965ts}, \cite{Ross:1983}) have proved that any Gittins search sequence against $\mathbf{p}$ is a pure strategy that optimally counters it.
This result is also recognized by a comment by Kelly on \cite{Gittins79}, which formulates the search game with known $\mathbf{p}$ as a multi-armed bandit problem optimally solved by Gittins indices from which the sequences in $\mathcal{C}_{\mathbf{p}}$ take their name.
Further, the proof of \cite{Ross:1983} shows the reverse is also true; in other words, any pure strategy optimally countering $\mathbf{p}$ must be a Gittins search sequence against $\mathbf{p}$.
Therefore, we say a Gittins search sequence against $\mathbf{p}$ is an \emph{optimal counter} to $\mathbf{p}$, and hence $\mathcal{C}_{\mathbf{p}}$ is the set of search sequences that optimally counter $\mathbf{p}$.

For any $\xi \in \mathcal{C}$, as a function of $\mathbf{p}$, $\sum_{i=1}^n p_i V_i(\xi)$ is a hyperplane in $n$-dimensional space. Combined with \eqref{eq:hider}, it follows that $v^*$ is the maximum of the lower envelope of an uncountable set of hyperplanes, which is a concave function of $\mathbf{p}$. Let $\mathcal{P}^*$ be the set of $\mathbf{p}$ attaining this maximum, so $\mathcal{P}^*$ is the set of optimal hiding strategies. In most cases, $|\mathcal{P}^*|=1$ and so the optimal hiding strategy is unique; an example with $|\mathcal{P}^*|>1$ can be found in Example \ref{example:p*notunique} in Section \ref{sec:properties}.

Since the hider maximizes a lower envelope of hyperplanes, there must exist at least one hyperplane containing the point $(\mathbf{p}^*,v^*)$ for all $\mathbf{p}^* \in \mathcal{P}^*$.
In other words, there exists at least one search sequence which optimally counters every optimal hiding strategy. 
The following proposition states that an optimal search strategy, if it exists, can only mix such search sequences.

\begin{proposition} \label{prop:searcheroptcounter}
Any optimal search strategy $\eta^*$ is a mixture of some subset of $\cap_{\mathbf{p} \in \mathcal{P}^*} \mathcal{C}_{\mathbf{p}}$.
\end{proposition}
\begin{proof}
Since the search game $G$ has a value $v^*$, if the searcher chooses $\eta^*$ and the hider any $\mathbf{p}^*\equiv(p_1^*,\ldots , p_n^*) \in \mathcal{P}^*$, the expected time until detection is $v^*$; in other words, we have
\begin{equation} \label{eqn:searcheroptv*}
v^* = \sum_{\xi \in \mathcal{C}} \eta^*(\xi) \left(\sum_{i=1}^n p_i^* V_i(\xi)\right).
\end{equation}
Suppose the statement of the proposition is false, so there exist $\bar{\mathbf{p}}\equiv(\bar{p}_1, \ldots , \bar{p}_n) \in \mathcal{P}^*$ and $\bar{\xi} \notin \mathcal{C}_{\bar{\mathbf{p}}}$ such that $\eta^*(\bar{\xi})>0$.
Since $\bar{\mathbf{p}}$ is optimal for the hider, $\sum_{i=1}^n \bar{p}_i V_i(\xi)\geq v^*$ for all $\xi \in \mathcal{C}$.
Because $\bar{\xi} \notin \mathcal{C}_{\bar{\mathbf{p}}}$, however, $\sum_{i=1}^n \bar{p}_i V_i(\bar{\xi})>v^*$.
Consequently, the right-hand side of \eqref{eqn:searcheroptv*} must be strictly greater than $v^*$ when $\mathbf{p}^*=\bar{\mathbf{p}}$, leading to a contradiction.
\end{proof}


It is intuitive that adding a new box will increase the value of the game, because the hider has one more place to hide, so the searcher needs to cover more ground, as seen in the next proposition.

\begin{proposition} \label{prop:pstar>0}
The following statements are true.
\begin{enumerate} [label={[\arabic*.]}]
\item If $\mathbf{p}^*\equiv(p_1^*,\ldots , p_n^*)$ is optimal for the hider, then $p_i^*>0$ for $i = 1, \ldots ,n$.
\item If $\eta^*$ is optimal for the searcher, then $V_i(\eta^*) = v^*$ for $i = 1,\ldots,n$.
\item Adding a new box increases the value of the game.
\end{enumerate}
\end{proposition}

\begin{proof}
We begin by proving [1.], which concerns the hider.
Write $v(\mathbf{p})$ for the expected time to detection when the hider chooses $\mathbf{p}$ and the searcher chooses any search sequence in $\mathcal{C}_{\mathbf{p}}$; therefore, any optimal hiding strategy maximizes $v(\mathbf{p})$. Let $\xi_{\mathbf{p}}$ be the element of $\mathcal{C}_{\mathbf{p}}$ which, when multiple boxes satisfy \eqref{eq:simpleGI}, searches the box with the smallest label.


To prove the statement by contradiction, suppose that we have $\mathbf{p}^*\equiv(p^*_1,\ldots,p^*_n) \in \mathcal{P}^*$ with $p^*_k=0$ for some $k \in \{ 1, \ldots, n\}$. Without loss of generality, relabel the boxes such that we have $p^*_n=0$, so
\begin{equation} \label{eqn:vP^*}
v(\mathbf{p}^*)=\sum_{i=1}^{n-1} p^*_i V_i(\xi_{\mathbf{p}^*}).
\end{equation}
Take any $\epsilon>0$ and consider $\mathbf{\bar{p}}\equiv(\bar{p}_1,\ldots , \bar{p}_n)$, where
\begin{align*}
\bar{p}_i&= p^*_i(1-\epsilon), \qquad i= 1, \ldots , n-1; \\
\bar{p}_n&= \epsilon.
\end{align*}

Compare $\xi_{\mathbf{\bar{p}}} \in \mathcal{C}_{\mathbf{\bar{p}}}$ and $\xi_{\mathbf{p}^*} \in \mathcal{C}_{\mathbf{p}^*}$. 
Both apply the same rule when multiple boxes satisfy \eqref{eq:simpleGI}, and for any $i,j \in \{1, \ldots , n-1\}$, we have
$$\frac{\bar{p}_i}{\bar{p}_j}=\frac{p^*_i(1-\epsilon)}{p^*_j(1-\epsilon)}=\frac{p^*_i}{p^*_j}.$$
Therefore, the subsequence of $\xi_{\mathbf{\bar{p}}}$ consisting of searches of boxes $1,2,\ldots, n-1$  is identical to $\xi_{\mathbf{p}^*}$.
In other words, $\xi_{\mathbf{\bar{p}}}$ is just $\xi_{\mathbf{p}^*}$ with searches of box $n$ inserted between some searches of the first $n-1$ boxes. 
Hence, we must have $V_i(\xi_{\mathbf{\bar{p}}})>V_i(\xi_{\mathbf{p}^*})$ for any $i \in \{1, \ldots, n-1\}$ with $p^*_i>0$. 
Further, we may choose $\epsilon$ small enough so that $\xi_{\mathbf{\bar{p}}}$ does not search box $n$ until at least $v(\mathbf{p}^*)$ time units have passed, ensuring $V_n(\xi_{\mathbf{\bar{p}}})>v(\mathbf{p}^*)$. From these observations and \eqref{eqn:vP^*}, it follows that
\begin{align*} 
v(\mathbf{\bar{p}})&=\epsilon V_n(\xi_{\mathbf{\bar{p}}})+ \sum_{i=1}^{n-1} p^*_i(1-\epsilon)V_i(\xi_{\mathbf{\bar{p}}}) \\
&> \epsilon v(\mathbf{p}^*) + (1-\epsilon) \sum_{i=1}^{n-1} p^*_iV_i(\xi_{\mathbf{p}^*})= v(\mathbf{p}^*),
\end{align*}
contradicting the optimality of $\mathbf{p}^*$, and therefore proving [1.].

Next, we prove [2.], concerning the searcher.
Suppose $\eta^*$ is optimal for the searcher and $\mathbf{p}^*$ is optimal for the hider. The expected time to detection under the strategy pair $(\mathbf{p}^*,\eta^*)$ is
\begin{equation} \label{eqn:v^*equaliseproof}
v^*=\sum_{i=1}^n p_i^*V_i(\eta^*).
\end{equation}
To prove by contradiction, suppose that $V_j(\eta^*) < v^*$ for some $j \in \{1,\ldots,n\}$. By \eqref{eqn:v^*equaliseproof}, there must either exist $k \in \{1,\ldots,n\}$ such that $V_k(\eta^*) > v^*$, or we must have $p_j^*=0$. The former cannot happen as $\eta^*$ guarantees the searcher an expected time to detection of at most $v^*$. The latter cannot happen by 1., 
which leads to a contradiction proving [2.].

Finally, we prove [3.] by showing that $v_{n+1}^* > v_n^*$, where $v_n^*$ is the value of an $n$-box game, and $v_{n+1}^*$ is the value if a new box is added to the $n$-box game.
In the game with $n+1$ boxes, the hider can guarantee an expected payoff of at least $v_n^*$ by not hiding in the new box, so $v_{n+1}^* \geq v_n^*$. However, any such strategy has $p_{n+1}=0$ so is not optimal by [1.].
Therefore, $v_n^*$ is not the value of the $(n+1)$-box game, so $v_{n+1}^* > v_n^*$, proving [3.].
\end{proof}

Note that [1.] in Proposition \ref{prop:pstar>0} is also proved by \cite{Bram1963} for unit-search-time $G$ via a different method to the proof above.


%

\subsection{An Equivalent Search Game}
\label{sec:altsearchgame}

In this section, we first show that an optimal search strategy exists in a modified version of the search game $G$. We then draw the same conclusion for $G$ by showing that any optimal search strategy in the modified game is also optimal in $G$.

Consider a search game $G(\epsilon)$, parametrized by $\epsilon \in (0, 1/n)$, identical to $G$ in all aspects apart from the set of pure search strategies, which are constructed by the following.
For $i=1,\ldots,n$, write
\begin{equation} \label{eqn:Mi}
M_i(\epsilon) \equiv \inf \{V_i(\xi): \xi \in \mathcal{C}_{\mathbf{p}} \; \; \text{with} \; \; p_i<\epsilon \}.
\end{equation}
In words, among all Gittins search sequences against hiding strategies with $p_i < \epsilon$, $M_i(\epsilon)$ is the smallest expected time to detection if the hider is in box $i$.
Unlike $G$, in $G(\epsilon)$, a pure strategy $\zeta_i (\epsilon)$ is available to the searcher for $i=1, \ldots, n$.
When selected, for $i = 1, \ldots, n$, $\zeta_i(\epsilon)$ results in payoff $M_i(\epsilon)$ if the hider is in box $i$ or payoff 0 otherwise.
In addition, available to the searcher in $G(\epsilon)$ are Gittins search sequences against hiding strategies in $\mathcal{P}(\epsilon)$, where
\begin{equation*} 
\mathcal{P}(\epsilon) \equiv \{\mathbf{p}: p_i \geq \epsilon, \; i = 1, \ldots , n\}.
\end{equation*}
To summarize, in $G(\epsilon)$, the searcher has the following pure strategy set:
\begin{equation} \label{eqn:SstarG}
\mathcal{C}(\epsilon) \equiv \{\xi \in \mathcal{C}_{\mathbf{p}}: \mathbf{p}\in \mathcal{P}(\epsilon) \} \cup \{ \zeta_i(\epsilon): i = 1, \ldots , n\}.
\end{equation}

For any $\epsilon \in (0, 1/n)$, since the payoff in $G(\epsilon)$ is bounded below by 0, by the standard results for semi-finite games (see, for example, Chapter 13 in \cite{Ferguson2020}), we can conclude that $G(\epsilon)$ has a value and optimal hiding strategy. 
To prove that $G(\epsilon)$ has an optimal search strategy, we consider its $\mathcal{S}$-game formulation (see start of Section \ref{sec:OSSexist}), in which the searcher chooses a vector in
$$\mathcal{S}(\epsilon) \equiv \{(V_1(\xi),\ldots, V_n(\xi)): \xi \in \mathcal{C}(\epsilon)\} \subset \mathbb{R}^n.$$
The following lemma is proven in Appendix \ref{append:closedlemma} by showing that $\mathcal{S}(\epsilon)$ is closed and applying Theorem 2.4.2 of \cite{Blackwell1954}.


\begin{lemma} \label{lemma:altminimaxbandhider}
For any $\epsilon \in (0,1/n)$, the game $G(\epsilon)$ has an optimal search strategy which is a mixture of at most $n$ search sequences.
\end{lemma}

%
%
%


The following result draws upon the properties in Section \ref{sec:propsforproof} to conclude that, for small enough $\epsilon$, the games $G$ and $G(\epsilon)$ are almost equivalent.

\begin{lemma} \label{lem:equivgames}
Consider $G$ and its set of optimal hiding strategies $\mathcal{P}^*$.
For any $\mathbf{p}^* \in \mathcal{P}^*$, there exists $\epsilon_{\mathbf{p}^*} \in (0,1/n]$ such that, for all $\epsilon \in (0, \epsilon_{\mathbf{p}^*})$, the games $G$ and $G(\epsilon)$ share the same value, $\mathbf{p}^*$ is optimal in $G(\epsilon)$ as well as in $G$, and a search strategy is optimal in $G$ if and only if it is optimal in $G(\epsilon)$.
\end{lemma}

\begin{proof}
Let $\mathbf{p}^* \equiv (p_1^*, \ldots , p_n^*) \in \mathcal{P}^*$ and $\epsilon_1 \equiv \min_{i\in \{1, \ldots , n\}} p_i^*$; we have $\epsilon_1>0$ by [1.] in Proposition \ref{prop:pstar>0}. 
Further, under any mixed hiding strategy, some box is chosen with at most probability $1/n$, so $\epsilon_1 \leq 1/n$.


The function $M_i(\epsilon)$ in \eqref{eqn:Mi} decreases in $\epsilon$, since the set over which the infimum is taken grows with $\epsilon$.
Write $\mathbf{p}\equiv (p_1, \ldots, p_n)$. If $p_i=0$, then any $\xi \in \mathcal{C}_{\mathbf{p}}$ never searches box $i$, so $V_i(\xi)$ is infinite, and hence $M_i(\epsilon) \uparrow \infty$ as $\epsilon \downarrow 0$.
On the other hand, if $p_i=1$, then any $\xi \in \mathcal{C}_{\mathbf{p}}$ only searches box $i$, so $V_i(\xi)=t_i/q_i$, and hence $M_i(\epsilon) \downarrow t_i/q_i \leq v^*$ as $\epsilon \uparrow 1$, where $v^*$ is the value of $G$.
Combining the above information, we may conclude that
$$\epsilon_2 \equiv \sup\{\epsilon: M_i(\epsilon) > v^*/p_i^*, \; i=1,\ldots , n \}$$
exists, and $M_i(\epsilon) > v^*/p_i^*$ for all $\epsilon \in (0,\epsilon_2)$, $i=1,\ldots , n$.

Let $\epsilon_{\mathbf{p}^*} \equiv \min(\epsilon_1,\epsilon_2)$; we show that $\epsilon_{\mathbf{p}^*}$ satisfies the conditions of the lemma. 
For any $\bar{\mathcal{C}} \subset \mathcal{C}$, write
\[
v(\mathbf{p},\bar{\mathcal{C}}) \equiv \inf_{\xi \in \bar{\mathcal{C}}} v(\mathbf{p}, \xi),
\]
where $v(\mathbf{p}, \xi)$ is the expected time to detection if the hider chooses $\mathbf{p}$ and the searcher chooses $\xi$.
Throughout the following, let $\epsilon \in (0, \epsilon_{\mathbf{p}^*})$.

Recall from \eqref{eqn:SstarG} that $\mathcal{C}(\epsilon)$ is the pure strategy set in $G(\epsilon)$.
In $G$, a hiding strategy $\mathbf{p}$ is optimally countered by any sequence in $\mathcal{C}_{\mathbf{p}}$, leading to an expected time to detection of $v(\mathbf{p},\mathcal{C})$, where $\mathcal{C}$ is the pure strategy set in $G$.
Bearing the above in mind, we show that $v(\mathbf{p},\mathcal{C}) \geq v(\mathbf{p},\mathcal{C}(\epsilon))$ for any hiding strategy $\mathbf{p}$ by considering two cases.
\begin{enumerate}
\item $\mathbf{p} \in \mathcal{P}(\epsilon)$. In this case, $\mathcal{C}_{\mathbf{p}}$ is contained in $\mathcal{C}(\epsilon)$; therefore, if the hider chooses $\mathbf{p}$, the searcher does no worse when $\mathcal{C}$ is replaced with $\mathcal{C}(\epsilon)$, so $v(\mathbf{p},\mathcal{C}) \geq v(\mathbf{p},\mathcal{C}(\epsilon))$.

\item $\mathbf{p} \notin \mathcal{P}(\epsilon)$. In this case, there exists $j \in \{1, \ldots , n\}$ such that $p_j<\epsilon$. By the construction of $\zeta_j(\epsilon)$, we have $V_i(\zeta_j(\epsilon)) \leq V_i(\xi)$ for any $\xi \in \mathcal{C}_{\mathbf{p}}$, $i = 1, \ldots , n$. Therefore, $\zeta_j(\epsilon)\in \mathcal{C}(\epsilon) \setminus \mathcal{C}$ dominates any sequence in $\mathcal{C}_{\mathbf{p}}$. 
It follows that $v(\mathbf{p},\mathcal{C}) \geq v(\mathbf{p},\mathcal{C}(\epsilon))$. 
\end{enumerate}

Now consider $v(\mathbf{p}^*, \mathcal{C}(\epsilon))$; by the above, $v^*=v(\mathbf{p}^*,\mathcal{C}) \geq v(\mathbf{p}^*,\mathcal{C}(\epsilon))$. Since $\epsilon < \epsilon_1$, we have $\mathbf{p}^* \in \mathcal{P}(\epsilon)$, and hence $\mathcal{C}_{\mathbf{p}^*} \subset \mathcal{C}(\epsilon)$. The only pure search strategies in $\mathcal{C}(\epsilon)$ that, when the hider chooses $\mathbf{p}^*$, could achieve a lower expected time to detection than a sequence in $\mathcal{C}_{\mathbf{p}^*}$ are those not in $\mathcal{C}$, namely $\{\zeta_i(\epsilon), \; i= 1, \ldots , n\}$. Therefore,  we have
\begin{align*}
v(\mathbf{p}^*,\mathcal{C}(\epsilon))&=\min \left(v^*, \min_{i \in \{1, \ldots, n\}} \left\{ \sum_{j=1}^n p^*_j V_j(\zeta_i(\epsilon)) \right\} \right) \\
&=\min \left(v^*, \min_{i \in \{1, \ldots, n\}} p^*_iM_i(\epsilon) \right) = v^*,
\end{align*}
where the final equality holds since $\epsilon < \epsilon_2$.

To conclude, we have $v(\mathbf{p}, \mathcal{C}) \geq v(\mathbf{p}, \mathcal{C}(\epsilon))$ for all hiding strategies $\mathbf{p}$, and $v^*=v(\mathbf{p}^*, \mathcal{C})=v(\mathbf{p}^*, \mathcal{C}(\epsilon))$. It follows that $\mathbf{p}^*$ is optimal in $G(\epsilon)$ and the value of $G(\epsilon)$ is $v^*$.




As for the searcher, since the set of pure hiding strategies and the value are the same for $G$ and $G(\epsilon)$, any optimal search strategy in $G$ is optimal in $G(\epsilon)$ if it is available to the searcher in $G(\epsilon)$ and vice versa. 
By Proposition \ref{prop:searcheroptcounter},
any optimal search strategy in $G$ chooses only search sequences in $\mathcal{C}_{\mathbf{p}^*}$, available in $G(\epsilon)$ since $\mathcal{C}_{\mathbf{p}^*} \subset \mathcal{C}(\epsilon)$. Further, for $i=1,\ldots , n$, we have $p^*_iM_i(\epsilon)>v^*$. Therefore, it is suboptimal in $G(\epsilon)$ for the searcher to choose any strategy in $\mathcal{C}(\epsilon) \setminus \mathcal{C} = \{\zeta_i(\epsilon),\; i=1,\ldots , n\}$. It follows that a search strategy is optimal in $G$ if and only if it is optimal in $G(\epsilon)$, completing the proof.
\end{proof}

We conclude this section with its main result.

\begin{theorem} \label{thm:OSSexist}
In the search game $G$, for any optimal hiding strategy $\mathbf{p}^*$, there exists an optimal search strategy which is a mixture of at most $n$ elements of $\mathcal{C}_{\mathbf{p}^*}$.
\end{theorem}
\begin{proof}
By Lemma \ref{lemma:altminimaxbandhider}, there exists a search strategy $\eta^*$, optimal in $G(\epsilon)$, which is a mixture of at most $n$ search sequences. By Lemma \ref{lem:equivgames}, $\eta^*$ is also optimal in the search game $G$. By Proposition \ref{prop:searcheroptcounter}, for any optimal hiding strategy $\mathbf{p}^*$, the search sequences mixed by $\eta^*$ must all belong to $\mathcal{C}_{\mathbf{p}^*}$, completing the proof.
\end{proof}


\section{Properties of Optimal Strategies} \label{sec:properties}
While we have shown that each player has an optimal strategy in the search game $G$, it turns out that each player's optimal strategy need not be unique.
In this section, we demonstrate how to identify an optimal hider-searcher strategy pair, present an example where the hider has multiple optimal strategies, and show that the searcher may always choose a simple optimal strategy among the many available. These findings will underpin the development of an efficient algorithm to compute a solution to $G$ in Section~\ref{sec:findopt}.

We begin by combining Propositions \ref{prop:searcheroptcounter} and \ref{prop:pstar>0} to identify simple conditions on a hider-searcher strategy pair which are both necessary and sufficient for optimality.

\begin{theorem} \label{thm:Git8.3}
Let $v$ be the expected time to detection when the hider chooses some mixed strategy $\mathbf{p}$ and the searcher some mixed strategy $\eta$. 
The mixed strategy $\mathbf{p}$ (resp. $\eta$) is optimal for the hider (resp. searcher) 
if and only if
\begin{enumerate}[label={[\Alph*.]}]
\item $\eta$ is a mixture of some subset of $\mathcal{C}_{\mathbf{p}}$. 
\item $V_i(\eta) = v$, for $i =1,\ldots, n$.
\end{enumerate}
\end{theorem}
\begin{proof}
First, we prove the forwards implication. If $\mathbf{p}$ is optimal for the hider and $\eta$ is optimal for the searcher, then [A.] follows from Proposition \ref{prop:searcheroptcounter}. 
Further, if $\mathbf{p}$ and $\eta$ are optimal, then $v$ is the value of the game, so [B.] follows from [2.] in Proposition \ref{prop:pstar>0}.

Second, we prove the backwards implication. By [A.], $\mathbf{p}$ guarantees an expected time to detection of at least $v$ regardless of what the searcher does. In addition, by [B.], $\eta$ guarantees an expected time to detection equal to $v$ regardless of what the hider does. Therefore, neither the hider nor the searcher can obtain a better guarantee than $v$; it follows that $\mathbf{p}$ and $\eta$ are an optimal strategy pair.
\end{proof}

Note that the backwards implication of Theorem \ref{thm:Git8.3} is Theorem 8.3 of \cite{Gittins:1989} applied to the search game.


Theorem \ref{thm:Git8.3} shows that any optimal search strategy $\eta^*$ is an \emph{equalizing} strategy; in other words, whenever the searcher plays $\eta^*$, the expected time to detection is the same no matter where the hider hides. Theorem 5.2 of \cite{Ruckle:1991} shows that, in the search game with unit search times, there exists an equalizing pure search strategy $\xi$. 
However, $\xi$ is not necessarily optimal by Theorem \ref{thm:Git8.3} because it is not necessarily a Gittins search sequence against any hiding strategy.
Further, the proof of Theorem 5.2 of \cite{Ruckle:1991} does not extend to the search game with arbitrary search times, as it relies on $V_k(\xi)$ being unaffected when the positions of a search of box $i$ and box $j$ in $\xi$ are switched.

We next use Theorem~\ref{thm:Git8.3} to demonstrate that it is possible for the hider to have multiple optimal strategies.

\begin{example} \label{example:p*notunique}
Consider a two-box search game where box $i$ has search time $t_i$ and detection probability $q_i$, $i=1,2$, with $q_1 < q_2$ and $t_1>t_2$.
Write $p$ for the probability that the hider hides in box 1. Inspection of \eqref{eq:simpleGI} shows that if
\begin{equation} \label{eqn:optpinexample}
p \in \left[\frac{q_2/t_2}{q_2/t_2+q_1/t_1}, \frac{q_2/t_2}{q_2/t_2+q_1(1-q_1)/t_1} \right]  ,
\end{equation}
then there exists a Gittins search sequence against $p$ that begins by searching box 1, followed by box 2, and then box 1 again.

Suppose that 
\begin{equation} \label{eqn:cycliccondexample}
(1-q_2)=(1-q_1)^2;
\end{equation} 
therefore, for any $p$, if the searcher makes, in any order, two unsuccessful searches of box 1 and one unsuccessful search of box 2, then the  posterior probability that the hider is in box 1 returns to $p$, and hence the problem has reset itself. 
It follows that the sequence $\xi$ that repeats the cycle of boxes $1,2,1$ indefinitely is a Gittins search sequence against any $p$ satisfying \eqref{eqn:optpinexample}.

Calculate
\begin{align*}
V_1(\xi) &= \sum_{k=1}^\infty (1-q_1)^{2(k-1)}q_1 \left[(k-1)(2t_1 + t_2) + t_1  + (1-q_1)k (2t_1+t_2)\right], \\
V_2(\xi) &= \sum_{k=1}^\infty (1-q_2)^{k-1} q_2 \left[(k-1)(2t_1 + t_2) + t_1+t_2\right].
\end{align*}
By rewriting
\begin{equation*}
V_1(\xi) = \sum_{k=1}^\infty (1-q_1)^{2(k-1)}[q_1 t_1+q_1(1-q_1)(2t_1+t_2)+(1-(1-q_1)^2)(k-1)x],
\end{equation*}
and using \eqref{eqn:cycliccondexample}, we have
$$V_1(\xi)-V_2(\xi) = \left[q_1(t_1+(1-q_1)(2t_1+t_2)) - q_2(t_1+t_2) \right] \sum_{k=1}^\infty (1-q_2)^{k-1},$$
from which it follows that $V_1(\xi)=V_2(\xi)$ if and only if
\begin{equation} \label{eqn:exampleV1=V2condition}
q_1=\frac{t_1-t_2}{t_1},
\end{equation}
with the combination of \eqref{eqn:cycliccondexample} and \eqref{eqn:exampleV1=V2condition} determining $q_2$.
Under \eqref{eqn:cycliccondexample} and \eqref{eqn:exampleV1=V2condition}, by Theorem \ref{thm:Git8.3}, it follows that $\xi$ is optimal for the searcher and any $p$ satisfying \eqref{eqn:optpinexample} is optimal for the hider.

For a numerical example, if $q_1=0.4$, $q_2=0.64$, $t_1=1$ and $t_2=0.6$, then any $p \in [8/11,40/49]$ is optimal for the hider.
\end{example}

\bigskip


By Theorem~\ref{thm:Git8.3}, it is sufficient for the searcher to consider Gittins search sequences against any optimal hiding strategy. Therefore, if there exists an optimal $\mathbf{p}$ against which there is a unique Gittins search sequence $\xi$ (so $|\mathcal{C}_{\mathbf{p}}|=1$), then the pure strategy $\xi$ is optimal for the searcher. One example can be found in Example \ref{example:p*notunique}; aside from the two endpoints, any $p$ satisfying \eqref{eqn:optpinexample} is both optimal for the hider and has $\mathcal{C}_p=\{\xi\}$, so the pure strategy $\xi$ is optimal for the searcher. 

Interestingly, the condition $|\mathcal{C}_{\mathbf{p}}|=1$ for some optimal $\mathbf{p}$---a rare situation---is not always necessary for the existence of an optimal pure search strategy. \cite{Ruckle:1991} shows that in the search game with $q_i=0.5$ and $t_i=1$, $i=1,\ldots,n$, the unique optimal hiding strategy $\mathbf{p}^*$ selects each box with probability $1/n$, and therefore any search sequence repeatedly passing through the $n$ locations is in $\mathcal{C}_{\mathbf{p}^*}$, meaning $\mathcal{C}_{\mathbf{p}^*}$ is of infinite size. \cite{Ruckle:1991} further shows that one such search sequence is optimal, namely
$$1,2,\ldots,n,n,n-1, \ldots,1, n,n-1,\ldots, 1, \ldots, $$
which passes through the locations once in ascending order, then in descending order ad infinitum.
Since all boxes are identical, by symmetry, any sequence beginning by permuting $1,2,\ldots,n$ before applying the reverse permutation ad infinitum is also optimal. 

However, in most cases where $|\mathcal{C}_{\mathbf{p}}|>1$ for all optimal $\mathbf{p}$, any optimal search strategy is a mixed strategy, creating a more challenging case on which we focus henceforth.

By Definition \ref{def:GIP}, the next box searched by any Gittins search sequence against a hiding strategy $\mathbf{p}$ must satisfy \eqref{eq:simpleGI}. If $|\mathcal{C}_{\mathbf{p}}|>1$, at some point in the search, the searcher must encounter a \emph{tie} where some $k \in \{2,\ldots , n\}$ boxes satisfy \eqref{eq:simpleGI}. At such a tie, any Gittins search sequence must search the $k$ tied boxes next in some arbitrary order.
Thereafter, any two Gittins search sequences will be identical until another tie is encountered. Therefore, elements of $\mathcal{C}_{\mathbf{p}}$ differ from one another only in how they break ties.

How the searcher chooses to break ties is important, since knowledge of the searcher's tie-breaking preferences could be used by the hider to their advantage. Therefore, a mixed tie-breaking strategy is required by the searcher.
Write $S_n$ for the set of permutations of $\{1,\ldots , n\}$; a permutation $\sigma \in S_n$ serves as a preference ordering to choose which box to search next if a tie is encountered. For example, a tie between boxes 1, 2 and 4 is broken in the order 2, 1, 4 by the permutation $(2,3,1,4) \in S_4$.

Of initial interest are Gittins search sequences that break every tie using the same preference ordering.
Write $\xi_{\sigma,\mathbf{p}}$ for the Gittins search sequence against $\mathbf{p}$ that breaks every tie encountered using $\sigma \in S_n$. We define the following subset of $\mathcal{C}_{\mathbf{p}}$:
\begin{equation*}
\widehat{\mathcal{C}}_{\mathbf{p}} \equiv \{\xi_{\sigma,\mathbf{p}} : \sigma \in S_n\}.
\end{equation*}
Whilst $\mathcal{C}_{\mathbf{p}}$ could be an infinite set, $|\widehat{\mathcal{C}}_{\mathbf{p}}| \leq n!$ since $|S_n|=n!$. By Theorem \ref{thm:OSSexist}, there exists an optimal search strategy that is a mixture of at most $n$ elements of $\mathcal{C}_{\mathbf{p}^*}$ for any optimal hiding strategy $\mathbf{p^*}$. The aim of the remainder of this section is to show that the same holds true if we replace $\mathcal{C}_{\mathbf{p}^*}$ with $\widehat{\mathcal{C}}_{\mathbf{p}^*}$. 


For any search strategy $\eta$ (pure or mixed), write $V(\eta) \equiv (V_1(\eta),\ldots, V_n(\eta))$. For any hiding strategy $\mathbf{p}$, write
\begin{equation*} 
\mathcal{S}_{\mathbf{p}} \equiv \{V(\xi): \xi \in \mathcal{C}_{\mathbf{p}}\}\quad \text{and} \quad \widehat{\mathcal{S}}_{\mathbf{p}} \equiv \{V(\xi): \xi \in \widehat{\mathcal{C}}_{\mathbf{p}}\},
\end{equation*}
noting that $\widehat{\mathcal{S}}_{\mathbf{p}} \subset \mathcal{S}_{\mathbf{p}} \subset \mathbb{R}^n$.
Clearly, if some search strategy $\eta$ is a mixture of a subset of $\mathcal{C}_{\mathbf{p}}$, then $V(\eta)$ can be written as a convex combination of the elements in $\mathcal{S}_{\mathbf{p}}$. The following lemma
shows that the same statement is true replacing $\mathcal{S}_{\mathbf{p}}$ with $\widehat{\mathcal{S}}_{\mathbf{p}}$ if $\mathbf{p}$ hides in each box with a nonzero probability.


\begin{lemma} \label{lemma:convexhull}
For any hiding strategy $\mathbf{p}\equiv(p_1,\ldots,p_n)$ with $p_i > 0$, $i=1,\ldots , n$, the convex hull of $\mathcal{S}_{\mathbf{p}}$ is equal to the convex hull of $\widehat{\mathcal{S}}_{\mathbf{p}}$.
\end{lemma}

\begin{proof}
If $|\mathcal{C}_{\mathbf{p}}|=1$ then $\mathcal{S}_{\mathbf{p}}=\widehat{\mathcal{S}}_{\mathbf{p}}$ and the result is trivially true. For the rest of the proof, assume $|\mathcal{C}_{\mathbf{p}}|>1$.
 
First, we show that $\mathcal{S}_{\mathbf{p}}$ is compact. Since $\mathcal{S}_{\mathbf{p}} \subset \mathbb{R}^n$, by the Heine-Borel theorem, $\mathcal{S}_{\mathbf{p}}$ is compact if and only if it is both closed and bounded. 
By Lemma \ref{lemma:fixedpclosed} in Appendix \ref{append:closedlemma}, $\mathcal{S}_{\mathbf{p}}$ is closed.
To show $\mathcal{S}_{\mathbf{p}}$ is bounded, write 
$$v(\mathbf{p}, \xi) \equiv \sum_{i=1}^n p_i V_i(\xi),$$ the expected time to detection if the hider uses $\mathbf{p}$ and the searcher uses $\xi$. 
Consider the search sequence $\xi_1$ that repeats the cycle of searches $(1,2,\ldots , n)$ indefinitely.
Clearly $V_i(\xi_1)$ is finite for $i=1,\ldots , n$, so $v(\mathbf{p}, \xi_1)$ is also finite. Since $\mathcal{C}_{\mathbf{p}}$ is the set of optimal counters to $\mathbf{p}$, we must have $v(\mathbf{p}, \xi) \leq v(\mathbf{p}, \xi_1)$ for any $\xi \in \mathcal{C}_{\mathbf{p}}$. Since $p_i>0$, 
we must have $V_i(\xi)$ finite for $i=1,\ldots, n$ and any $\xi \in \mathcal{C}_{\mathbf{p}}$; it follows that $\mathcal{S}_{\mathbf{p}}$ is bounded and hence compact.

Write $\text{Conv}(\mathcal{S}_{\mathbf{p}})$ for the convex hull of $\mathcal{S}_{\mathbf{p}}$. By definition, $\text{Conv}(\mathcal{S}_{\mathbf{p}})$ is convex, and, since $\mathcal{S}_{\mathbf{p}}$ is compact in the finite-dimensional vector space $\mathbb{R}^n$, $\text{Conv}(\mathcal{S}_{\mathbf{p}})$ is also compact. (See Corollary 5.33 of \cite{Chara:2013}.)
Therefore, we may apply the Krein-Milman theorem to deduce that $\text{Conv}(\mathcal{S}_{\mathbf{p}})$ is equal to the convex hull of its extreme points. To prove Lemma \ref{lemma:convexhull}, we show that $\widehat{\mathcal{S}}_{\mathbf{p}}$ is the set of extreme points of $\text{Conv}(\mathcal{S}_{\mathbf{p}})$. We first note the following useful facts.

\begin{itemize}
\item By definition, a point $\mathbf{x} \in \text{Conv}(\mathcal{S}_{\mathbf{p}})$ is extreme if and only if, for any $\mathbf{y}, \mathbf{z} \in \text{Conv}(\mathcal{S}_{\mathbf{p}})$ and $\lambda \in (0,1)$ satisfying $\mathbf{x} = \lambda \mathbf{y} + (1-\lambda) \mathbf{z}$, we have $\mathbf{x} = \mathbf{y} = \mathbf{z}$. In other words, the only way we can express $\mathbf{x}$ as a convex combination of elements of $\text{Conv}(\mathcal{S}_{\mathbf{p}})$ is by $\mathbf{x}$ itself. 

\item 
For any $\mathbf{s}\equiv(s_1, \ldots , s_n) \in \mathcal{S}_{\mathbf{p}}$, the weighted average $\sum_{i=1}^n s_i p_i$ is equal to the expected time to detection if the hider chooses $\mathbf{p}$ and the searcher any optimal counter $\xi \in \mathcal{C}_{\mathbf{p}}$. Therefore, all elements of $\mathcal{S}_{\mathbf{p}}$ lie on the same hyperplane, say $H$, in $\mathbb{R}^n$.

\item By the definition of $\mathcal{S}_{\mathbf{p}}$, we have 
$$\mathcal{S}_{\mathbf{p}} \subset R \equiv \left\{(v_1, \ldots , v_n) \in \mathbb{R}^n: \min_{\xi \in \mathcal{C}_{\mathbf{p}}} V_i(\xi) \leq v_i \leq \max_{\xi \in \mathcal{C}_{\mathbf{p}}} V_i(\xi),\; i=1,\ldots , n \right\},$$
where $R$ is a hyperrectangle in $n$-dimensional space (also known as an $n$-orthotope).
Further, since $\text{Conv}(\mathcal{S}_{\mathbf{p}})$ is the smallest convex set containing $\mathcal{S}_{\mathbf{p}}$ and $R$ is also a convex set containing $\mathcal{S}_{\mathbf{p}}$, we have $\text{Conv}(\mathcal{S}_{\mathbf{p}}) \subseteq R$.

\end{itemize}


The proof will be done by double inclusion.
In the first half of the  double inclusion proof,
we show that any point in $\widehat{\mathcal{S}}_{\mathbf{p}}$ is an extreme point of $\text{Conv}(\mathcal{S}_{\mathbf{p}})$. 
Let $\mathbf{x}\equiv(x_1, \ldots , x_n) \in \widehat{\mathcal{S}}_{\mathbf{p}}$. Then $\mathbf{x}$ corresponds to a search sequence $\xi_{\sigma} \in \widehat{\mathcal{C}}_{\mathbf{p}}$ which breaks all ties using some $\sigma \in S_n$. 
Without a loss of generality, let $\sigma = (1,2\ldots, n)$. Suppose that
\begin{equation} \label{eqn:convexcombo}
\mathbf{x} = \lambda \mathbf{y} + (1-\lambda) \mathbf{z}
\end{equation}
for some $\mathbf{y}, \mathbf{z} \in \text{Conv}(\mathcal{S}_{\mathbf{p}})$ and $\lambda \in (0,1)$. To prove that $\mathbf{x}$ is extreme, we show that we must have $\mathbf{x}=\mathbf{y}=\mathbf{z}$. 

Since, when breaking any tie, $\xi_{\sigma}$ gives preference to box 1 over any other box, no other search sequence in $\mathcal{C}_{\mathbf{p}}$ makes the $j$th search of box 1 any sooner than $\xi_{\sigma}$, $j =1,2,\ldots $; therefore, we have $x_1 = \min_{\xi \in \mathcal{C}_{\mathbf{p}}} V_1(\xi)$. For any $\mathbf{v}\equiv(v_1,\ldots , v_n) \in \text{Conv}(\mathcal{S}_{\mathbf{p}})$, since $\text{Conv}(\mathcal{S}_{\mathbf{p}}) \subset R$, we must have $v_1 \geq x_1$. It follows that, for \eqref{eqn:convexcombo} to hold, we must have $y_1=z_1=x_1$.

We now demonstrate how this argument may be repeated to show that $y_2=z_2=x_2$. Let $\mathcal{C}_{1,\mathbf{p}}$ be the elements of $\mathcal{C}_{\mathbf{p}}$ which, when breaking any tie involving box 1, give preference to box 1; therefore, for any $\xi \in \mathcal{C}_{\mathbf{p}}$, we have $V_1(\xi)=x_1$ if and only if $\xi \in \mathcal{C}_{1,\mathbf{p}}$.
Write $V(\xi)\equiv (V_1(\xi), \ldots , V_n(\xi))$ and $\mathcal{S}_{1,\mathbf{p}} \equiv \{ V(\xi): \xi \in \mathcal{C}_{1,\mathbf{p}} \}$; therefore, for any $\mathbf{v}=(v_1,\ldots , v_n) \in \mathcal{S}_{\mathbf{p}}$, we have $v_1=x_1$ if and only if $\mathbf{v}\in \mathcal{S}_{1,\mathbf{p}}$.
Since $\mathcal{S}_{1,\mathbf{p}} \subset \mathcal{S}_{\mathbf{p}}$, we have $\text{Conv}(\mathcal{S}_{1,\mathbf{p}}) \subset \text{Conv}(\mathcal{S}_{\mathbf{p}})$. Note that $\mathbf{y}, \mathbf{z} \in \text{Conv}(\mathcal{S}_{1,\mathbf{p}})$ since 
$y_1=z_1=x_1$. 

Write
$$R_1 \equiv \left\{(x_1, v_2 ,\ldots , v_n) \in \mathbb{R}^n: \min_{\xi \in \mathcal{C}_{1,\mathbf{p}}} V_i(\xi) \leq v_i \leq \max_{\xi \in \mathcal{C}_{1,\mathbf{p}}} V_i(\xi),\; i=2,\ldots , n \right\}.$$
Then $R_1$, a hyperrectangle in $n-1$-dimensional space, is a convex set containing $\mathcal{S}_{1,\mathbf{p}}$, so $\text{Conv}(\mathcal{S}_{1,\mathbf{p}}) \subset R_1$. Since, when breaking any tie, $\xi_{\sigma}$ gives preference to box 1 over box 2, but then to box 2 over any other box $i$, $i = 3, \ldots , n$, no other sequence in $\mathcal{C}_{1,\mathbf{p}}$ makes the $j$th search of box 2 sooner than $\xi_{\sigma}$, $j =1,2,\ldots$; therefore, we have $x_2 = \min_{\xi \in \mathcal{C}_{1,\mathbf{p}}} V_2(\xi)$. Since any $\mathbf{v}\equiv(v_1,\ldots , v_n)\in \text{Conv}(\mathcal{S}_{1,\mathbf{p}})$ also belongs to $R_1$, we must have $v_2 \geq x_2$. It follows that, for \eqref{eqn:convexcombo} to hold, we must have $y_2=z_2=x_2$.

We may repeat the above argument a further $n-3$ times to conclude that $y_i=z_i=x_i$ for $i = 1,\ldots , n-1$. Finally, since $\mathbf{y}, \mathbf{z}$ and $\mathbf{x}$ all lie in the same hyperplane $H$ in $\mathbb{R}^n$, we must have $y_n=z_n=x_n$, so $\mathbf{y} = \mathbf{z} = \mathbf{x}$, and $\mathbf{x}$ must be an extreme point of $\text{Conv}(\mathcal{S}_{\mathbf{p}})$.

In the second half of the double inclusion proof, 
we show that any extreme point of $\text{Conv}(\mathcal{S}_{\mathbf{p}})$ is in $\widehat{\mathcal{S}}_{\mathbf{p}}$. We will prove the contrapositive of this statement; i.e., we show that any $\mathbf{a} \in \text{Conv}(\mathcal{S}_{\mathbf{p}}) \setminus \widehat{\mathcal{S}}_{\mathbf{p}}$ is not an extreme point of $\text{Conv}(\mathcal{S}_\mathbf{p})$.


To begin, note that, by definition, any element of $\text{Conv}(\mathcal{S}_{\mathbf{p}})$ can be written as a convex combination of some $m \in \{1,2,\ldots\}$ elements of $\mathcal{S}_{\mathbf{p}}$. If $\mathbf{b} \in \text{Conv}(\mathcal{S}_{\mathbf{p}}) \setminus \mathcal{S}_{\mathbf{p}}$, then any such convex combination must contain at least $m \geq 2$ elements of $\mathcal{S}_{\mathbf{p}}$, so $\mathbf{b}$ is not an extreme point of $\text{Conv}(\mathcal{S}_{\mathbf{p}})$.
Hence, our task is reduced to showing that any $\mathbf{a} \in \mathcal{S}_{\mathbf{p}} \setminus \widehat{\mathcal{S}}_{\mathbf{p}}$ is not an extreme point of $\text{Conv}(\mathcal{S}_{\mathbf{p}})$.

Consider $\mathbf{a} \in \mathcal{S}_{\mathbf{p}} \setminus \widehat{\mathcal{S}}_{\mathbf{p}}$, and write $\xi_{\mathbf{a}}$ for the corresponding search sequence in $\mathcal{C}_{\mathbf{p}}  \setminus \widehat{\mathcal{C}}_{\mathbf{p}}$ satisfying $\mathbf{a}=V(\xi_{\mathbf{a}})$. 
Since $\xi_{\mathbf{a}} \notin \widehat{\mathcal{C}}_{\mathbf{p}}$, there exists no permutation in $S_n$ with which $\xi_{\mathbf{a}}$ breaks every tie it encounters. It follows that there must exist some $k \in \{2, \ldots, n\}$ boxes (without loss of generality boxes $1,\ldots,k$) and $k$ ties encountered by $\xi_{\mathbf{a}}$ such that
no permutation of $\{1,\ldots,k\}$ serves as a preference ordering for how all $k$ ties are broken.
Suppose, again with no loss of generality, tie $m$ involves (at least) boxes $m$ and $m+1$, with box $m$ searched before box $m+1$ by $\xi_{\mathbf{a}}$, $m=1, \ldots, k-1$, and tie $k$ involves (at least) boxes $k$ and $1$, with box $k$ searched before box 1 by $\xi_{\mathbf{a}}$. 
Note that ties $1,\ldots , k$ are not necessarily consecutive nor in chronological order.



We now aim to construct a mixture of elements of $\mathcal{C}_{\mathbf{p}}$ which mimics the performance of $\xi_{\mathbf{a}}$; it will follow that $\mathbf{a}$ is not an extreme point of $\text{Conv}(\mathcal{S}_\mathbf{p})$.

Consider tie $k$, which is broken by $\xi_{\mathbf{a}}$ using the preference ordering
\[
j_{1},\ldots , j_{\alpha-1}, k, j_{\alpha+1}, \ldots , j_{\beta-1}, 1, j_{\beta+1}, \ldots, j_{n},
\]
bearing in mind that box $j_i$ is not necessarily involved in tie $k$, $i \in \{1,\ldots,n\}\setminus \{\alpha, \beta\}$.
In other words, box $k$ ranks in the $\alpha$th position, and box 1 ranks in the $\beta$th position, for some $1 \leq \alpha < \beta \leq n$.



Let $\xi_{\mathbf{a},\alpha} \in \mathcal{C}_{\mathbf{p}}$ break tie $k$ using preference ordering 
$$j_{1},\ldots , j_{\alpha-1}, 1,k, j_{\alpha+1}, \ldots , j_{\beta-1}, j_{\beta+1}, \ldots, j_{n},$$
and all other ties in the same order as $\xi_{\mathbf{a}}$. Similarly, let $\xi_{\mathbf{a},\beta} \in \mathcal{C}_{\mathbf{p}}$ break tie $k$ using preference ordering  
$$j_{1},\ldots , j_{\alpha-1}, j_{\alpha+1}, \ldots , j_{\beta-1}, 1, k, j_{\beta+1}, \ldots, j_{n},$$
and all other ties in the same order as $\xi_{\mathbf{a}}$. In other words, at tie $k$, both $\xi_{\mathbf{a},\alpha}$ and $\xi_{\mathbf{a},\beta}$ switch the order boxes 1 and $k$ are searched by $\xi_{\mathbf{a}}$ to prefer box 1, and all boxes searched between boxes $k$ and $1$ by $\xi_{\mathbf{a}}$ when breaking tie $k$ have (retaining their order) been shifted after boxes 1 and $k$ are searched in $\xi_{\mathbf{a},\alpha}$, and before boxes 1 and $k$ are searched in $\xi_{\mathbf{a},\beta}$. Note that if $\beta = \alpha+1$ (so there are no boxes searched between boxes $k$ and 1 by $\xi_{\mathbf{a}}$ when breaking tie $k$), then $\xi_{\mathbf{a},\alpha}=\xi_{\mathbf{a},\beta}$, but the following argument is still valid.

Note that $V_i(\xi_{\mathbf{a},\alpha})=V_i(\xi_{\mathbf{a},\beta})=V_i(\xi_{\mathbf{a}})$ for any box $i \in \{j_1, \ldots, j_{\alpha-1}, j_{\beta+1}, \ldots , j_n\}$. Recall $t_i$ is the search time of box $i$, and let
$$\eta_k \equiv \frac{t_k}{t_1+t_k}\xi_{\mathbf{a},\alpha} \oplus \frac{t_1}{t_1+t_k}\xi_{\mathbf{a},\beta}.$$
Clearly $V_i(\eta_k)=V_i(\xi_{\mathbf{a}})$ for $i \in \{j_1, \ldots, j_{\alpha-1}, j_{\beta+1}, \ldots , j_n\}$. For $i \in \{j_{\alpha+1}, \ldots , j_{\beta-1}\}$, let $w_i$ be the probability that the hider is found on the first search of box $i$ after the $k$th tie is reached, conditional on the hider being in box $i$. Then we have $V_i(\xi_{\mathbf{a},\alpha})=V_i(\xi_{\mathbf{a}})+w_it_1$ and $V_i(\xi_{\mathbf{a},\beta})=V_i(\xi_{\mathbf{a}})-w_it_k$ for $i \in \{j_{\alpha+1}, \ldots , j_{\beta-1}\}$. It follows that
$$V_i(\eta_k)= \frac{t_k}{t_1+t_k}\left(V_i(\xi_\mathbf{a})+w_it_1\right) + \frac{t_1}{t_1+t_k}\left(V_i(\xi_\mathbf{a})-w_it_k\right) = V_i(\xi_\mathbf{a})$$
for $i \in \{j_{\alpha+1}, \ldots , j_{\beta-1}\}$.

Because $V_k(\xi_{\mathbf{a},\beta}) > V_k(\xi_{\mathbf{a},\alpha})>V_k(\xi_{\mathbf{a}})$, we have $V_k(\eta_k)>V_k(\xi_{\mathbf{a}})$.
Since $V(\eta_k)$ and $V(\xi_{\mathbf{a}})$ lie in the same hyperplane, $H$, in $\mathbb{R}^n$, we must have $V_1(\eta_k)<V_1(\xi_{\mathbf{a}})$. To summarize, we have
\begin{equation} \label{eqn:etak}
V_k(\eta_k)>V_k(\xi_{\mathbf{a}}), \quad  V_1(\eta_k)<V_1(\xi_{\mathbf{a}}), \quad \text{and} \quad V_i(\eta_k)=V_i(\xi_{\mathbf{a}}) \; \text{for} \; i\neq 1, k.
\end{equation}

Now, for $m \in \{1,\ldots , k-1\}$, we repeat the same procedure with tie $m$ for boxes $m$ and $m+1$ to create $\eta_m$ which satisfies:
\begin{equation} \label{eqn:etaj}
V_m(\eta_m)>V_m(\xi_{\mathbf{a}}), \quad  V_{m+1}(\eta_m)<V_{m+1}(\xi_{\mathbf{a}}), \quad \text{and} \quad V_i(\eta_m)=V_i(\xi_{\mathbf{a}}) \; \; \text{for} \; \; i\neq m, m+1.
\end{equation}

To complete the proof, we develop a mixture of $\{\eta_1, \ldots , \eta_k\}$ which mimics the performance of $\xi_{\mathbf{a}}$. To begin, by \eqref{eqn:etaj} with $m=1,2$, for any $\lambda \in (0,1)$, the mixture
\begin{equation} \label{eqn:eta12mix}
\eta_{1,2}(\lambda) \equiv \lambda \eta_1 \oplus (1-\lambda) \eta_2
\end{equation}
satisfies 
\begin{equation*}
V_1(\eta_{1,2}(\lambda))>V_1(\xi_{\mathbf{a}}), \quad V_3(\eta_{1,2}(\lambda))<V_3(\xi_{\mathbf{a}}), \quad \text{and} \quad V_i(\eta_{1,2}(\lambda))=V_i(\xi_{\mathbf{a}}) \; \; \text{for} \; i=4, \ldots , n. 
\end{equation*}
Also by \eqref{eqn:etaj}, there exists $\lambda^* \in (0,1)$ such that $\eta_{1,2} \equiv \eta_{1,2}(\lambda^*)$ satisfies $V_2(\eta_{1,2})=V_2(\xi_{\mathbf{a}})$. Therefore, we have  
\begin{equation} \label{eqn:eta12}
V_1(\eta_{1,2})>V_1(\xi_{\mathbf{a}}), \quad V_3(\eta_{1,2})<V_3(\xi_{\mathbf{a}}), \quad \text{and} \quad V_i(\eta_{1,2})=V_i(\xi_{\mathbf{a}}) \; \; \text{for} \; \; i \neq 1,3. 
\end{equation}

By \eqref{eqn:etaj} with $m=3$ and \eqref{eqn:eta12}, there exists a mixture, $\eta_{1,2,3}$, of $\eta_{3}$ and $\eta_{1,2}$ satisfying
\begin{equation*} 
V_1(\eta_{1,2,3})>V_1(\xi_{\mathbf{a}}), \quad V_4(\eta_{1,2,3})<V_4(\xi_{\mathbf{a}}), \quad \text{and} \quad V_i(\eta_{1,2,3})=V_i(\xi_{\mathbf{a}}) \; \; \text{for} \; \; i\neq 1,4. 
\end{equation*}
We may repeat this process of mixing $\eta_{m}$ and $\eta_{1,\ldots, m-1}$ to create $\eta_{1,\ldots, m}$ for $m=4, \ldots, k-1$, with the resulting $\eta_{1,\ldots, k-1}$ satisfying
\begin{equation} \label{eqn:eta1..k-1}
V_1(\eta_{1,\ldots, k-1})>V_1(\xi_{\mathbf{a}}), \quad V_k(\eta_{1,\ldots, k-1})<V_k(\xi_{\mathbf{a}}), \quad \text{and} \quad V_i(\eta_{1,\ldots, k-1})=V_i(\xi_{\mathbf{a}}) \; \text{for} \; i\neq 1,k. 
\end{equation}
Finally, by \eqref{eqn:etak} and \eqref{eqn:eta1..k-1}, we may mix $\eta_k$ and $\eta_{1,\ldots , k-1}$ to create $\eta_{1,\ldots , k}$ satisfying $V_i(\eta_{1,\ldots , k})=V_i(\xi_{\mathbf{a}})$ for $i=2,\ldots , n$. Yet, since $V(\eta_{1,\ldots , k})$ and $V(\eta_k)$ both lie in $H$, we must also have $V_1(\eta_{1,\ldots , k})=V_1(\xi_{\mathbf{a}})$. It follows that, for some $\bar{\lambda} \in (0,1)$, we have 
$$\mathbf{a} =  V(\xi_{\mathbf{a}})= V(\eta_{1,\ldots , k})=\bar{\lambda} V(\eta_{1,\ldots , k}) + (1-\bar{\lambda}) V(\eta_{k}).$$
By the construction of the $\eta_m$ for $m = 1, \ldots , k$ as mixtures of elements of $\mathcal{C}_{\mathbf{p}}$, $V(\eta_{1,\ldots , k})$ and $V(\eta_{k})$ are two distinct elements in $\text{Conv}(\mathcal{S}_\mathbf{p})$ different from $\mathbf{a}$, showing that $\mathbf{a}$ is not an extreme point of $\text{Conv}(\mathcal{S}_\mathbf{p})$ and completing the proof.
\end{proof}

Lemma \ref{lemma:convexhull} allows us to present the main theorem of this section, which strengthens Theorem \ref{thm:OSSexist} by replacing $\mathcal{C}_{\mathbf{p}^*}$ with $\widehat{\mathcal{C}}_{\mathbf{p}^*}$.

\begin{theorem} \label{thm:OSSperm}
In search game $G$, for any optimal hiding strategy $\mathbf{p}^*$, there exists an optimal search strategy which is a mixture of at most $n$ elements of $\widehat{\mathcal{C}}_{\mathbf{p}^*}$.
\end{theorem}

\begin{proof}
Let $\mathbf{p}^* \equiv (p_1^*,\ldots , p_n^*)$ be an optimal hiding strategy. By Theorem \ref{thm:OSSexist}, there exists an optimal search strategy $\eta^*$ which is a mixture of elements of $\mathcal{C}_{\mathbf{p}^*}$. Therefore, $V(\eta^*)$ can be written as a convex combination of elements of $\mathcal{S}_{\mathbf{p}^*}$ and hence belongs to $\text{Conv}(\mathcal{S}_{\mathbf{p}^*})$, the convex hull of $\mathcal{S}_{\mathbf{p}^*}$.

By [1.] in Proposition~\ref{prop:pstar>0}, $p_i^*>0$ for $i = 1,\ldots , n$. Therefore, by Lemma \ref{lemma:convexhull}, $\text{Conv}(\mathcal{S}_{\mathbf{p}^*})=\text{Conv}(\widehat{\mathcal{S}}_{\mathbf{p}^*})$. It follows that $V(\eta^*) \in \text{Conv}(\widehat{\mathcal{S}}_{\mathbf{p}^*})$ and hence can be written as a convex combination of elements of $\widehat{\mathcal{S}}_{\mathbf{p}^*}$. 


By Carathéodory's theorem, $V(\eta^*)$ can be written as a convex combination of at most $n+1$ elements in $\widehat{\mathcal{S}}_{\mathbf{p}^*}$. Yet, for any hiding strategy $\mathbf{p} \equiv(p_1,\ldots,p_n)$ and $\mathbf{s} \equiv (s_1, \ldots , s_n) \in \mathcal{S}_{\mathbf{p}}$, the weighted average $\sum_{i=1}^n s_i p_i$ is equal to the expected time to detection if the hider chooses $\mathbf{p}$ and the searcher any optimal counter $\xi \in \mathcal{C}_{\mathbf{p}}$. Therefore, all elements of $\mathcal{S}_{\mathbf{p}}$ lie on the same hyperplane in $\mathbb{R}^n$, and hence so do all elements of $\widehat{\mathcal{S}}_{\mathbf{p}} \subset \mathcal{S}_{\mathbf{p}}$. It follows that the number of elements in the convex combination of $V(\eta^*)$ can be reduced to at most $n$, so $\eta^*$ is a mixture of at most $n$ strategies in $\widehat{\mathcal{C}}_{\mathbf{p}^*}$.
\end{proof}



Proposition 8.5 of \cite{Gittins:1989} proves a special case of Theorem \ref{thm:OSSperm} with $n=2$ and $t_1 = t_2 = 1$, but that proof does not extend directly to $n \geq 3$; see Appendix \ref{append:Git} for some discussion.
Our result applies to an arbitrary number of boxes and to general search times.
The significance of Theorem \ref{thm:OSSperm} is that, in order to construct an optimal search strategy, it is sufficient to consider Gittins search sequences which use the same preference ordering to break every tie.

\section{Computing Optimal Strategies} \label{sec:findopt}

This section uses the results developed in Sections \ref{sec:OSSexist} and \ref{sec:properties} to determine or estimate optimal strategies for both players.
Since the searcher's pure strategy space $\mathcal{C} \equiv \{1,2,\ldots , n\}^{\infty}$ is uncountable, the search game $G$ 
is a semi-finite game, whose optimal strategies are generally difficult to determine.
To make progress, we consider finite subgames of $G$.

Write $G_\mathcal{D}$ for the subgame of $G$ where the searcher's pure strategies are some finite $\mathcal{D} \subset \mathcal{C}$.
The subgame $G_\mathcal{D}$ is a finite, $n \times |\mathcal{D}|$ matrix game, which can be solved by linear programming (see \cite{Washburn:2003un}) with optimal strategies guaranteed to exist for both players. Write $v_{\mathcal{D}}^*$ for the value of $G_{\mathcal{D}}$. Since $\mathcal{D} \subset \mathcal{C}$, $v^*_{\mathcal{D}}$ is an upper bound on $v^*$, the value of $G$. We begin with an optimality test for a hiding strategy in $G$.

\begin{proposition} \label{prop:subgamesoln}
Consider $\mathbf{p} \equiv (p_1,\ldots , p_n)$ with $p_i>0$ for $i=1,\ldots, n$, and write $\mathcal{D} \equiv \widehat{\mathcal{C}}_{\mathbf{p}}$, the set of Gittins search sequences against $\mathbf{p}$ that use the same preference ordering to break every tie. Let $\eta$ be an optimal search strategy in the  $n \times |\mathcal{D}|$ matrix game $G_\mathcal{D}$, where $|\mathcal{D}| \in \{1,\ldots , n!\}$.
The following three statements are equivalent.
\begin{enumerate}[label=\roman*)]
\item $\mathbf{p}$ is optimal in $G$; \label{enu:i}
\item $\mathbf{p}$ is optimal in $G_\mathcal{D}$; \label{enu:ii}
\item $\eta$ is optimal in $G$. \label{enu:iii}
\end{enumerate}
\end{proposition}

\begin{proof}
First, we prove that \ref{enu:i} implies \ref{enu:ii}, so suppose $\mathbf{p}$ is optimal in $G$. By Theorem \ref{thm:OSSperm}, there exists a search strategy $\eta^*$ which is both (a) optimal in $G$, so guarantees the searcher an expected time to detection of at most $v^*$, and (b) a mixture of strategies in $\widehat{\mathcal{C}}_{\mathbf{p}}=\mathcal{D}$, so is available to the searcher in the game $G_{\mathcal{D}}$.
Since $\mathbf{p}$ is optimal in $G$, $\mathbf{p}$ guarantees the hider at least $v^*$ in $G$; since $\mathcal{D} \subset \mathcal{C}$, $\mathbf{p}$ has the same guarantee for the hider in $G_\mathcal{D}$. By (a) and (b) above, the searcher can guarantee at most $v^*$ with $\eta^*$ in $G_\mathcal{D}$. It follows that $\mathbf{p}$ is optimal in $G_\mathcal{D}$.


Second, we prove that \ref{enu:ii} implies \ref{enu:iii}, so suppose $\mathbf{p}$ is optimal in $G_\mathcal{D}$. Since $\eta$ is optimal in $G_\mathcal{D}$, then $\eta$ guarantees the searcher an expected time to detection of at most $v^*_{\mathcal{D}}$, no matter which box the hider hides in. Therefore, $V_i(\eta) \leq v^*_{\mathcal{D}}$ for $i=1, \ldots , n$. By the minimax theorem for finite games, when the hider plays $\mathbf{p}$ and the searcher plays $\eta$, the expected time to detection is $v^*_{\mathcal{D}}$. In other words, we have
$$\sum_{i=1}^n p_i V_i(\eta)= v^*_{\mathcal{D}}.$$
Since $p_i>0$ for $i=1,\ldots , n$, we must have $V_i(\eta) = v^*_{\mathcal{D}}$ for $i=1,\ldots , n$. In addition, $\eta$ is a mixture of strategies in $\mathcal{D} \subseteq \mathcal{C}_{\mathbf{p}}$. 
By Theorem \ref{thm:Git8.3}, $\eta$ is optimal in $G$.

Finally, we prove that \ref{enu:iii} implies \ref{enu:i}, so suppose $\eta$ is optimal in $G$. By Proposition \ref{prop:pstar>0}, $V_i(\eta)=v^*$ for $i=1,\ldots , n$. Further, because $\eta$ is available to the searcher in $G_\mathcal{D}$, it is a mixture of strategies in $\mathcal{D} \subseteq \mathcal{C}_{\mathbf{p}}$. By Theorem \ref{thm:Git8.3}, $\mathbf{p}$ is optimal in $G$, completing the proof.
\end{proof}

We say a hiding strategy $\mathbf{p} \equiv (p_1,\ldots , p_n)$ is \emph{interior} if $p_i>0$ for $i=1,\ldots, n$; otherwise, we say $\mathbf{p}$ is \emph{exterior}. By [1.] in Proposition \ref{prop:pstar>0} any optimal hiding strategy is interior, and we can test the optimality of any interior hiding strategy using Proposition~\ref{prop:subgamesoln}.
The hiding strategy $\mathbf{p}_0 \equiv (p_{0,1}, \ldots ,p_{0,n})$ with
\begin{equation} \label{eqn:p0formula}
p_{0,i} \equiv \frac{t_i/q_i}{\sum_{j=1}^n t_j/q_j}, \quad i=1,\ldots , n,
\end{equation}
is of particular interest, since it creates a tie between the Gittins indices of all $n$ boxes in \eqref{eq:simpleGI} at the start of the search, giving the searcher no preference over which box to search first. \cite{GitRob1} and \cite{GitRob2} both numerically find that $\mathbf{p}_0$ is optimal for the hider in many (but not all) unit-search-time problems. 
Further, the former proves $\mathbf{p}_0$ is optimal in a two-box problem with $(1-q_1)^m=(1-q_2)^{m+1}$ if and only if $m \leq 12$.
As discussed in Section \ref{sec:properties}, \cite{Ruckle:1991} solves the game with $n$ identical boxes and finds $\mathbf{p}_0$, which here hides in each box with probability $1/n$, to be optimal.
With Proposition~\ref{prop:subgamesoln}, we can quickly test whether $\mathbf{p}_0$ is optimal and also compute an optimal search strategy if so. Proposition \ref{prop:subgamesoln} is put into practice in the numerical experiments of Section \ref{sec:numerical}.

If $\mathbf{p}_0$ is suboptimal, Proposition \ref{prop:subgamesoln} is less useful for finding an optimal hiding strategy. For this instance, we develop an algorithm that estimates an optimal strategy for each player by successively computing tighter bounds on $v^*$.
Recall that, for any $\mathcal{D} \subset \mathcal{C}$, $v^*_\mathcal{D}$ is an upper bound on $v^*$. In addition, for any hiding strategy $\mathbf{p}$, $v(\mathbf{p})$ is a lower bound on $v^*$, where $v(\mathbf{p})$ is the expected time to detection when the hider chooses $\mathbf{p}$ and the searcher chooses any search sequence in $\mathcal{C}_{\mathbf{p}}$. Therefore, any finite set of search sequences and hiding strategy induce bounds on $v^*$, which is the main idea in the following algorithm.

\vspace{10pt}
\begin{algorithm}
\label{al:minExpCost}
\begin{enumerate}
\item \label{step:setup}  Set $L=0$ as a lower bound and $U=\infty$ as an upper bound for $v^*$, and pick $\epsilon > 0$ and a large integer $M$ so that the algorithm will stop either when $U/L-1 < \epsilon$, or after $M$ iterations.
Set $\mathbf{p}=\mathbf{p}_0$, and initialize $\mathcal{D}$ with the $n$ elements of $\widehat{\mathcal{C}}_{\mathbf{p}}$ which break all ties using preference orderings $(1,2,\ldots,n)$, $(2,3,\ldots,n, 1), \ldots$, $(n,1,2,\ldots,n-1)$.

\item \label{step:solveandcheck}
Solve the finite matrix game $G_\mathcal{D}$
and write $\mathbf{p}_\mathcal{D}$ for the corresponding optimal hiding strategy.
If $\mathbf{p}_\mathcal{D}$ is exterior, use Algorithm~\ref{al:get_interior} to add search sequences to $\mathcal{D}$ and solve $G_\mathcal{D}$ again; repeat until $\mathbf{p}_\mathcal{D}$ becomes interior.
Update $\mathbf{p} \leftarrow \mathbf{p}_\mathcal{D}$.

\item \label{step:update_bounds} Update $U \leftarrow v^*_\mathcal{D}$.
Calculate $v(\mathbf{p})$ and update $L \leftarrow \max(L, v(\mathbf{p}))$.

\item \label{step:check_stop}
If either $U/L-1 < \epsilon$, or the number of iterations reaches $M$, stop and output $v^*_\mathcal{D}$, $\mathbf{p}$, and the optimal search strategy for $G_\mathcal{D}$; otherwise, update $\mathcal{D} \leftarrow \mathcal{D} \cup \{\xi\}$ for any $\xi \in \mathcal{C}_{\mathbf{p}}$ and go to step \ref{step:solveandcheck}.

\end{enumerate}
\end{algorithm}

\begin{figure}[h!]
\begin{center}
\includegraphics[scale=0.55]{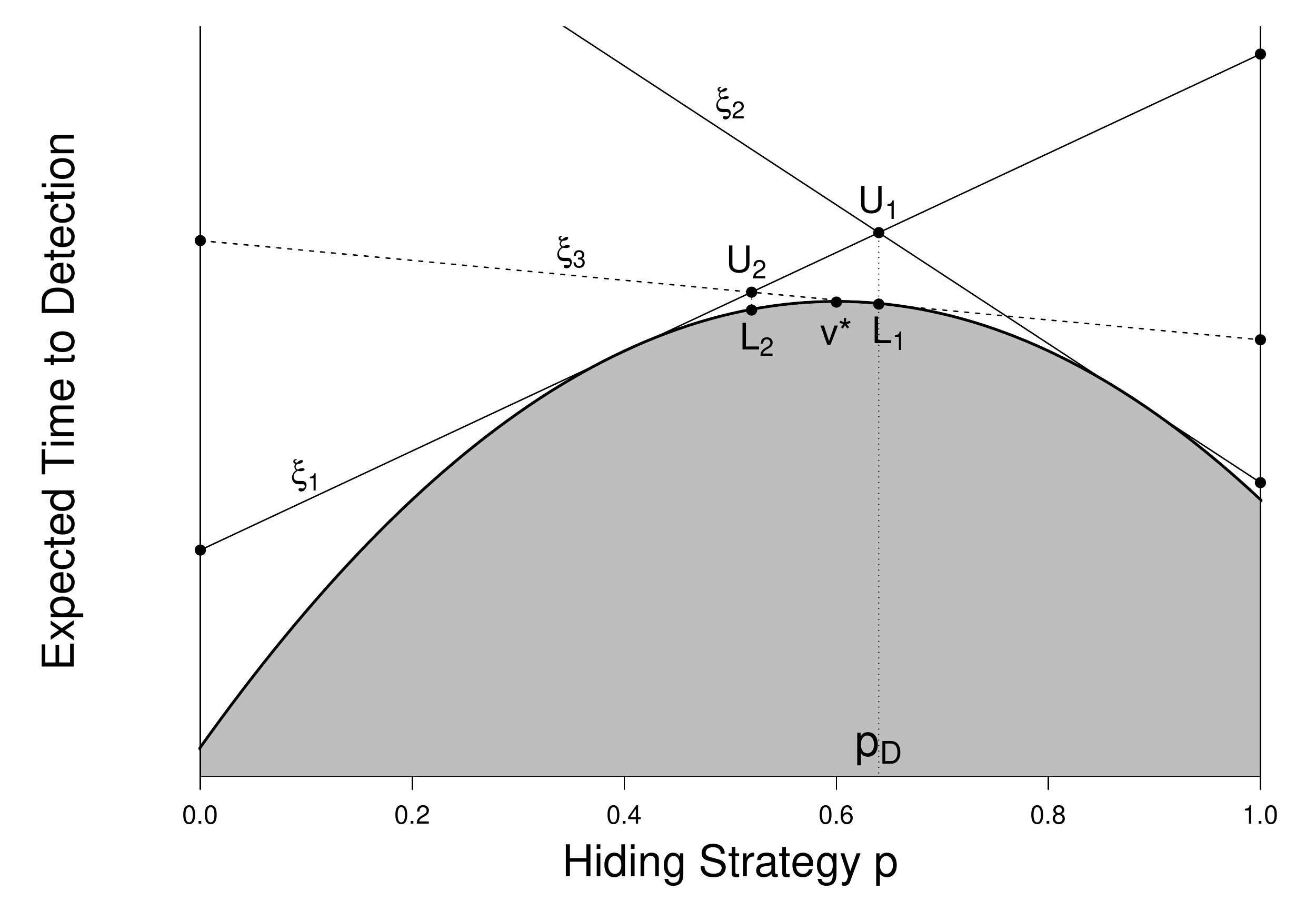}
\caption{Progressive calculation of upper and lower bounds. The solid tangent lines represent the set of search sequences in the current iteration, and the dashed tangent line is the new search sequence generated to be used in the next iteration.}
\label{fig:cuttingplane}
\end{center}
\end{figure}

The overall rationale of Algorithm \ref{al:minExpCost} is best understood via an example with $n=2$ boxes, where any mixed hiding strategy can be delineated by $p \in [0,1]$ which represents the probability of hiding in box 1.
Figure~\ref{fig:cuttingplane} demonstrates Algorithm \ref{al:minExpCost} in action. Each straight line represents the expected time to detection for a search sequence as the hiding strategy $p$ varies in $[0,1]$.
The function $v(p)$ is the lower envelope of the set of all search sequences, hence, a concave function in $p$, as indicated by the bold curve in Figure~\ref{fig:cuttingplane}.
We seek to determine $v^* \equiv \max_{p \in [0,1]} v(p)$.
Suppose $\mathcal{D}$ is initialized with two search sequences represented by the two solid straight lines $\xi_1$ and $\xi_2$.
By mixing these two search sequences, the searcher's optimal strategy in $G_\mathcal{D}$ guarantees that the expected time to detection is no more than $U_1 \equiv v^*_\mathcal{D}$, an upper bound for $v^*$.
The optimal hiding strategy in $G_\mathcal{D}$, namely $p_\mathcal{D}$, is used to generate a new search sequence $\xi_3$ (the dashed straight line), a Gittins search sequence against $p_\mathcal{D}$, with corresponding expected time to detection, $L_1$, a lower bound for $v^*$.
Furthermore, by adding $\xi_3$ to $\mathcal{D}$, in the next iteration we can compute a new, tighter upper bound $U_2$ by allowing the searcher to mix any subset of $\xi_1$, $\xi_2$ and $\xi_3$ (in Figure \ref{fig:cuttingplane}, the searcher mixes $\xi_1$ and $\xi_3$). A new lower bound, $L_2$, which may or may not be the overall best lower bound to date, is derived from the optimal hiding strategy in the new game $G_\mathcal{D}$.

For arbitrary $n$, the process is identical, but the straight lines representing search sequences are hyperplanes in $n$-dimensional space. In particular, for $n=3$, search sequences are represented by planes, whose lower envelope becomes a dome. 

Algorithm \ref{al:minExpCost} continues until either $U/L < 1+\epsilon$, or $M$ iterations have been completed. The final hiding strategy guarantees the hider an expected time to detection of at least $L$, the final search strategy guarantees the searcher an expected time to detection of at most $U$, and $v^*$ lies between $L$ and $U$.

Whilst the above paints an overall picture of how Algorithm \ref{al:minExpCost} works, the remainder of this section will explain the rationale behind each step of Algorithm \ref{al:minExpCost}.

\paragraph{Step \ref{step:setup}}
\cite{GitRob1} and \cite{GitRob2} find that when $\mathbf{p}_0$ in \eqref{eqn:p0formula} is not optimal, $v(\mathbf{p}_0)$ often approximates $v^*$ well.
Therefore, initializing $\mathbf{p}=\mathbf{p}_0$ and $\mathcal{D}$ with a subset of $\widehat{\mathcal{C}}_{\mathbf{p}_0}$ starts the algorithm with a lower bound $v(\mathbf{p})$ close to $v^*$.

For moderate to large $n$, it is not computationally feasible to initialize $\mathcal{D}$ with all elements of $\widehat{\mathcal{C}}_{\mathbf{p}_0}$, which could number up to $n!$. Further, even for small $n$, we found no numerical evidence to support initializing $\mathcal{D}$ with more than $n$ elements of $\widehat{\mathcal{C}}_{\mathbf{p}_0}$.
In Algorithm \ref{al:minExpCost}, we choose $n$ elements of $\widehat{\mathcal{C}}_{\mathbf{p}_0}$ by cycling the preference ordering $(1,\ldots, n)$ as described in step \ref{step:setup} to include a variety of tie-breaking strategies against $\mathbf{p}_0$

\paragraph{Step \ref{step:solveandcheck}}
We test if $\mathbf{p}_\mathcal{D}$ is exterior to ensure that $\mathbf{p}$ remains interior throughout Algorithm \ref{al:minExpCost} for the following reason.
If $\mathbf{p}$ in step \ref{step:update_bounds} has $p_j=0$ for some $j \in \{1,\ldots , n\}$, then any $\xi \in \mathcal{C}_{\mathbf{p}}$ will never search box $j$, so $V_j(\xi)$ will be infinite and $\xi$ is of no use to the searcher. Since $\xi$ is the only search sequence added to $\mathcal{D}$ in step \ref{step:check_stop}, the algorithm is prohibited from moving forward.

If $\mathbf{p}_\mathcal{D}$ is exterior, the following algorithm describes how we use the current interior hiding strategy $\mathbf{p}$ to add search sequences to $\mathcal{D}$ so $\mathbf{p}_\mathcal{D}$ becomes interior.  
\vspace{10pt}

\begin{algorithm} \label{al:get_interior}
\begin{enumerate}[label=2.(\alph*)]

\item \label{stepei:relabel} 

Write $p_{\mathcal{D},i}$ (resp. $p_i$) for the $i$th element of $\mathbf{p}_\mathcal{D}$ (resp.~$\mathbf{p}$), $i=1,\ldots,n$, and define $\mathcal{I} \equiv \{i : p_{\mathcal{D},i}=0\}$ and $\alpha \equiv \sum_{i \in \mathcal{I}} p_i>0$.


\item \label{stepei:perturb}
Obtain a new hiding strategy $\mathbf{\bar{p}}$ by setting
\begin{alignat*}{2}
\bar{p}_i &\leftarrow \be p_i, \quad \; &&i \in \mathcal{I} \\
\bar{p}_i &\leftarrow p_{\mathcal{D},i} (1-\be \alpha),  \quad \; &&i \in \{1,\ldots, n\} \setminus \mathcal{I},
\end{alignat*}
where $\beta \in (0,1)$ is a predetermined scaler.

\item \label{stepei:update}
Pick some $\xi \in \mathcal{C}_{\bar{\mathbf{p}}}$ arbitrarily and update $\mathcal{D} \leftarrow \mathcal{D} \cup \{\xi\}$.

\item \label{stepei:checkint}
Solve $G_{\mathcal{D}}$. 
If $\mathbf{p}_\mathcal{D}$ is interior, stop. Otherwise, update $\mathbf{p} \leftarrow \mathbf{\bar{p}}$ 
and go to step \ref{stepei:relabel}.
\end{enumerate}
\end{algorithm}

The rationale of Algorithm \ref{al:get_interior} can be understood as follows. If $i \in \mathcal{I}$, then $p_{\mathcal{D},i}=0$, so, with the current set of search sequences $\mathcal{D}$ available to the searcher, the hider does not want to hide in box $i$. To entice the hider into box $i$, 
the searcher needs to add a search sequence to $\mathcal{D}$ that searches in box $i$ 
less frequently 
than those currently in $\mathcal{D}$.


The last sequence added to $\mathcal{D}$ was a Gittins search sequence against $\mathbf{p}$. Therefore, in step \ref{stepei:update}, we add to $\mathcal{D}$ a Gittins search sequence $\xi$ against a hiding strategy $\bar{\mathbf{p}}$ (created in step \ref{stepei:perturb}) with $\bar{p}_i<p_i$, $i \in \mathcal{I}$, and ratio of hiding probabilities in boxes not in $\mathcal{I}$ the same as in $\mathbf{p}_\mathcal{D}$ (an optimal hiding strategy in the most recently-solved $G_\mathcal{D}$). 
Compared with any sequence in $\mathcal{C}_{\mathbf{p}}$, $\xi$ will search less frequently in any box $i \in \mathcal{I}$, and more frequently across boxes not in $\mathcal{I}$. 
The process may be repeated by reducing $\bar{p}_i$ further to eventually generate $\mathbf{p}_\mathcal{D}$ with $p_{\mathcal{D},i} > 0$, $i \in \mathcal{I}$.


\paragraph{Step \ref{step:update_bounds}}
Here, we discuss the updates to the bounds $L$ and $U$.
Note that if $\mathcal{D} \subseteq \mathcal{D}'$, then $v^*_\mathcal{D} \geq v^*_{\mathcal{D}'}$, since the searcher cannot do worse using $\mathcal{D}'$ than using $\mathcal{D}$.
Hence, since in both steps \ref{step:solveandcheck} and \ref{step:check_stop} we update $\mathcal{D}$ only by adding search sequences to it, the upper bound in each iteration will be at least as good as the one from the previous iteration, and thus the best upper bound to date.
In fact, the new upper bound will be strictly better than the previous upper bound, unless we have found $v^*$; see Proposition \ref{prop:UBimprove}.

It is, however, possible that the lower bound in the current iteration is worse than lower bounds in previous iterations (see Figure \ref{fig:cuttingplane} for a demonstration).
Hence, in step \ref{step:update_bounds}, we keep the best known lower bound to date.

\begin{proposition}
\label{prop:UBimprove}
In Algorithm~\ref{al:minExpCost}, write $U_k$ for the upper bound in step \ref{step:update_bounds} in iteration $k$, for $k=1,2,\ldots$.
If $U_k > v^*$, then $U_{k+1} < U_k$, for $k=1,2,\ldots$.
\end{proposition}
\begin{proof}
In iteration $k$ of Algorithm \ref{al:minExpCost}, write $\mathbf{p}_k$ for the interior hiding strategy updated in step \ref{step:solveandcheck}, $G_k$ for the subgame solved with $\mathbf{p}_k$ an optimal hiding strategy, $L_k$ for the lower bound in step \ref{step:update_bounds}, and $\xi_k \in \mathcal{C}_
{\mathbf{p}_k}$ for the new search sequence added to $\mathcal{D}$ in step \ref{step:check_stop}.
We already know that $U_{k+1} \leq U_k$; to prove the proposition by contradiction, we suppose that $U_{k+1} = U_k$ and show that $U_k = v^*$.

Since $U_{k+1} = U_k$ and the set of pure hiding strategies do not change from iteration $k$ to iteration $k+1$, the searcher must have a strategy optimal in $G_{k+1}$ that is available in $G_k$. Therefore, $\mathbf{p}_k$, optimal for the hider in $G_k$, is also optimal in $G_{k+1}$. It follows that $\mathbf{p}_k$ guarantees the hider an expected time to detection of at least $U_{k+1}$ regardless of the strategy of the searcher in $G_{k+1}$. Since $\xi_k \in \mathcal{C}_
{\mathbf{p}_k}$ is available to the searcher in $G_{k+1}$, we must have
\begin{equation}
U_{k+1} \leq v(\mathbf{p}_k).
\label{eq:C>U}
\end{equation}
On the other hand, by construction, we have
\begin{equation}
v(\mathbf{p}_k) \leq L_k.
\label{eq:L>C}
\end{equation}
Equations (\ref{eq:C>U}) and (\ref{eq:L>C}) imply that $U_k = U_{k+1} \leq v(\mathbf{p}_k) \leq L_{k}$.
Together with $L_k \leq v^* \leq U_k$, we can conclude that $L_k = v^* = U_k$, which completes the proof.
\end{proof}

\paragraph{Step \ref{step:check_stop}}
While Algorithm~\ref{al:minExpCost} attempts to tighten the upper bound and lower bound for $v^*$ through iterations, to guarantee that Algorithm~\ref{al:minExpCost} will terminate, we stop after a large, prespecified number of iterations $M$.
In the computational experiments of Section \ref{sec:numerical}, with $\epsilon=10^{-6}$ and $M=150$, we always obtained $U/L-1 < \epsilon$ with fewer than $M$ iterations; see Table \ref{tab:sseqsconveps} for more details on the number of iterations required to achieve convergence.

\section{Numerical Experiments} \label{sec:numerical}
This section presents several numerical experiments to demonstrate the efficiency of Algorithm \ref{al:minExpCost} and evaluate the performance of $\mathbf{p}_0$ defined in \eqref{eqn:p0formula} as a heuristic strategy for the hider.

In order to evaluate a Gittins search sequence 
which uses a particular tie-breaking rule, the searcher needs to properly recognize a tie between the Gittins indices in \eqref{eq:simpleGI}.
Comparing indices directly, however, does not yield reliable results because the indices are encoded as floating-point numbers. To overcome this obstacle, we design our numerical experiments to focus on two types of search games---\textit{cyclic} search games and \textit{acyclic} search games---and, for each, develop appropriate techniques to compute the conditional expected time to detection based on the hider's location under any Gittins search sequence.

For cyclic search games, there exist some coprime, positive integers $x_i$, $i=1,\ldots,n$, such that
\begin{equation} \label{eqn:cyclic}
(1-q_1)^{x_1}=\cdots = (1-q_n)^{x_n}.
\end{equation}
After $x_i$ searches of box $i$, for $i=1,\ldots,n$, the posterior probability vector on the hider's location returns to the initial $\mathbf{p}$, so the cyclic search game has reset itself.
For acyclic search games, for any distinct $i,j \in \{1,\ldots, n\}$, we require
\begin{equation} \label{eqn:acyclic}
(1-q_i)^{x}\neq (1-q_j)^{y}
\end{equation}
for any strictly positive integers $x$ and $y$.

\subsection{Calculating Expected Time to Detection} \label{sec:numercalcs}
The expected value of any nonnegative-valued random variable $X$ can be calculated by $E[X] = \int_0^\infty P\{X>x\} dx$. Using this formula, the expected time to detection if the searcher uses a search sequence $\xi$ and the hider hides in box $i$, for $i=1, \ldots,n$, can be calculated by
\begin{equation} \label{eqn:genv_i(xi)}
V_i(\xi) = \lim_{K \rightarrow \infty} \sum_{k=1}^{K} (1-q_i)^{k-1} [b_i(k,\xi)-b_i(k-1,\xi)],
\end{equation}
where $b_i(k,\xi)$ is the time at which the $k$th search of box $i$ is made under $\xi$. 
If $\xi$ is a Gittins search sequence against some hiding strategy $\mathbf{p}$, the terms $b_i(k,\xi)$ are determined by the Gittins indices in \eqref{eq:simpleGI} and the rule $\xi$ uses to break ties between these indices.
Comparing indices in \eqref{eq:simpleGI} directly, however, does not reliably recognize ties,
because detection probabilities, search times, and $\mathbf{p}$ must all be encoded as floating-point numbers. In this subsection, we describe methods to reliably calculate \eqref{eqn:genv_i(xi)} for cyclic and acyclic search games.

First, consider cyclic search games.
In step \ref{step:setup} of Algorithm \ref{al:minExpCost}, we need to evaluate a Gittins search sequence, $\xi_\sigma$, against $\mathbf{p}_0$ which breaks every tie using some preference ordering $\sigma$.
At the beginning of the search, all $n$ indices are tied, so the first $n$ searches of $\xi_\sigma$ will correspond to the order of $\sigma$.
Due to \eqref{eqn:cyclic}, after the first $n$ searches we can reliably compare indices and hence reliably recognize ties by keeping track of the number of searches that $\xi_\sigma$ has performed in each box---as opposed to comparing floating-point indices directly (see Appendix \ref{append:numerdetails} for details).
By \eqref{eqn:cyclic}, in the first $\sum_{i=1}^n x_i$ searches, any Gittins search sequence against $\mathbf{p}_0$ searches box $i$ exactly $x_i$ times, for $i=1,\ldots, n$. At that point, again by \eqref{eqn:cyclic}, all $n$ indices are tied, so the problem has reset itself.
Consequently, $\xi_\sigma$ will repeat the same cycle of $\sum_{i=1}^n x_i$ searches indefinitely, leading to a closed form for $V_i(\xi_\sigma)$, $i=1,\ldots,n$; see Appendix \ref{append:numerdetails} for details.

In step \ref{step:check_stop} of Algorithm \ref{al:minExpCost} and step \ref{stepei:update} of Algorithm \ref{al:get_interior}, we also need to evaluate some $\xi \in \mathcal{C}_{\mathbf{p}}$ where $\mathbf{p}$ is a solution to a finite matrix game. Yet, because we can take any $\xi \in \mathcal{C}_{\mathbf{p}}$ in these two steps, it is inconsequential whether ties between indices are recognized.
To calculate $V_i(\xi)$, note that after an initial transient period, the aforementioned cycle of $\sum_{i=1}^n x_i$ searches will repeat indefinitely, again leading to a closed form for $V_i(\xi)$; see Appendix \ref{append:numerdetails} for details.

Second, consider acyclic games.
When evaluating the Gittins search sequence $\xi_\sigma$ against $\mathbf{p}_0$ which breaks ties using some preference ordering $\sigma$, the first $n$ searches will correspond to the order of $\sigma$.
Due to \eqref{eqn:acyclic}, we will not encounter any ties 
from the $n$th search onwards, so we can calculate $\xi_\sigma$ by comparing floating-point indices directly.
The same technique can be used from the beginning of the search to evaluate $\xi \in \mathcal{C}_{\mathbf{p}}$ with $\mathbf{p}$ a solution to a finite matrix game.
To compute both $V_i(\xi_\sigma)$ and $V_i(\xi)$, first note that the partial sum of the first $K$ terms in \eqref{eqn:genv_i(xi)} provides a lower bound.
To obtain an upper bound, after the first $K$ searches in box $i$, adopt a search sequence that visits box $i$ at fixed intervals less frequently than any Gittins search sequence.
We increase $K$ until the ratio between the upper and lower bound is within $1 + 10^{-10}$.
See Appendix \ref{append:numerdetails} for details.

\subsection{Sample Schemes}
We now introduce the numerical study, beginning with the generation of boxes, for which acyclic and cyclic search games require different methods.
To generate a search time and detection probability for a box in an acyclic search game, we draw
\begin{equation} \label{eqn:acyclicdraw}
q_i \sim U(q_l,q_u), \quad t_i \sim U(1,5), \quad i=1,\ldots,n,
\end{equation}
for pre-specified $0<q_l<q_u<1$. The resulting search game with $n$ such boxes will be acyclic, since for any $q_i$ and $q_j$ drawn from a continuous uniform distribution, the event $\log(1-q_i)/\log(1-q_j) \in \mathbb{Q}$ has probability 0, and hence \eqref{eqn:acyclic} is satisfied almost surely.

To generate search times and detection probabilities for a set of $n$ boxes in a cyclic search game, we draw
\begin{equation} \label{eqn:cyclicdraw}
q_1 \sim U(q_l,q_u), \quad  x_i \sim DU(1,10), \quad t_i \sim U(1,5), \quad i=2,\dots , n,
\end{equation}
for pre-specified $0<q_l<q_u<1$, 
with $DU(1,10)$ representing the discrete uniform distribution where each integer in $\{1, \ldots , 10\}$ is selected with equal probability. 
Whilst $q_1$ is drawn directly by \eqref{eqn:cyclicdraw}, for $i = 2, \ldots , n$, we attain $q_i$ using $x_i$, $q_1$, and the relationship in \eqref{eqn:cyclic}.
To allow comparisons of acyclic and cyclic search games generated using the same $q_l$ and $q_u$, we use \eqref{eqn:cyclicdraw} with rejection sampling, rejecting a search game if the draws in \eqref{eqn:cyclicdraw} lead to $q_i \notin [q_l,q_u]$ for any $i \in \{2,\ldots , n\}$. 
 
We study four schemes based on different values of $q_l$ and $q_u$, as seen in Table \ref{tab:dpstvar}.
\begin{table}
\caption{Sample schemes used throughout the numerical study by values of $q_l$ and $q_u$ used in \eqref{eqn:acyclicdraw} and \eqref{eqn:cyclicdraw}. 
} \label{tab:dpstvar}
\begin{center}  
\begin{tabular}{|c| c|} 
 \hline
Sample Scheme & $[q_l,q_u]$ \\
\hline
Varied & $[0.1,0.9]$ \\
Low    & $[0.1,0.5]$ \\
Medium & $[0.3,0.7]$ \\
High   & $[0.5,0.9]$ \\
\hline
\end{tabular}
\end{center}
\end{table}
Search games with $n=2$, $3$, $5$ and $8$ boxes will be investigated. To account for increased variation within a search game as $n$ increases, for each value of $n$ and sample scheme, we study both $n \times 1000$ acyclic and $n \times 1000$ cyclic search games. Results are presented in the next subsection.

\subsection{Numerical Results} \label{sec:numerres}
For each generated search game, we first test the optimality of $\mathbf{p}_0$ using Proposition \ref{prop:subgamesoln}, which involves solving the finite game $G_\mathcal{D}$ where the searcher is restricted to the set of pure strategies $\mathcal{D} \equiv\widehat{\mathcal{C}}_{\mathbf{p}_0}$. By Proposition \ref{prop:subgamesoln}, $\mathbf{p}_0$ is optimal in $G$ if and only if $\mathbf{p}_0$ is optimal in $G_{\mathcal{D}}$. Since $G_{\mathcal{D}}$ may have multiple optimal hiding strategies, to determine the optimality of $\mathbf{p}_0$ in $G_{\mathcal{D}}$, we compare $v^*_\mathcal{D}$, the value of $G_\mathcal{D}$, to $v(\mathbf{p}_0)$, the expected time to detection when the hider plays $\mathbf{p}_0$ and the searcher plays any search sequence in $\mathcal{C}_{\mathbf{p}_0}$. In principle, $\mathbf{p}_0$ is optimal in $G_{\mathcal{D}}$ if and only if $v(\mathbf{p}_0)$ and $v^*_\mathcal{D}$ are equal. 
Due to limitations in computational accuracy, however, we accept equality if $|v(\mathbf{p}_0) - v^*_\mathcal{D}| / v^*_\mathcal{D} < 10^{-6}$. 
Table \ref{tab:results} presents the percentage of search games in which $\mathbf{p}_0$ is optimal for different sample schemes for $n=2, 3, 5$.
Since $|\widehat{\mathcal{C}}_{\mathbf{p}_0}|\in\{1,\ldots,n!\}$, it is often computationally infeasible to solve $G_{\mathcal{D}}$ and hence perform this test for $n=8$.

Next, for each search game where Proposition \ref{prop:subgamesoln} finds $\mathbf{p}_0$ to be suboptimal, we run Algorithm \ref{al:minExpCost} to estimate the value $v^*$ and optimal strategies. In step \ref{step:solveandcheck} of Algorithm \ref{al:minExpCost} and step \ref{stepei:checkint} of Algorithm \ref{al:get_interior}, we check whether $\mathbf{p}_\mathcal{D}$ is exterior. Whilst in principle $\mathbf{p}_\mathcal{D} \equiv (p_{\mathcal{D},1}, \ldots , p_{\mathcal{D},n})$ is exterior if and only if $\min_{i \in \{1,\ldots , n\}} p_{\mathcal{D},i} = 0$, due to limits in computational precision, we accept $\mathbf{p}_\mathcal{D}$ as exterior if $\min_{i \in \{1,\ldots , n\}} p_{\mathcal{D},i} < 10^{-6}$.

In step \ref{stepei:perturb} of Algorithm \ref{al:get_interior}, a predetermined scalar $\beta \in (0,1)$ determines a new hiding strategy $\bar{\mathbf{p}}$ such that some $\xi \in \mathcal{C}_{\bar{\mathbf{p}}}$ is added to the set of search sequences $\mathcal{D}$. 
The closer $\be$ is to 1, the more iterations that are required in Algorithm \ref{al:get_interior}, but the search sequence added to $\mathcal{D}$ during Algorithm \ref{al:get_interior} is more similar to a Gittins search sequence against $\mathbf{p}$, so fewer iterations are required for Algorithm~\ref{al:minExpCost}.
We find setting $\beta$ between 0.6 and 0.8 leads to the fastest convergence of Algorithm \ref{al:minExpCost}, so we set $\beta=0.7$ in all numerical experiments.



Recall that Algorithm~\ref{al:minExpCost} terminates either when the ratio of the upper and lower bound on $v^*$, namely $U/L$, is within $1+\epsilon$, or after a prespecified number of iterations $M$.
We set $M=150$ for all numerical tests and found Algorithm \ref{al:minExpCost} always terminated (with  $U/L<1+\epsilon$) with fewer than 150 iterations for $\epsilon=10^{-6}$ or greater.
Table \ref{tab:sseqsconveps} reports the mean and 95th percentile of the number of search sequences in the set $\mathcal{D}$ needed for the convergence of $L$ and $U$ with the \emph{varied} sample scheme for $\epsilon=10^{-3}, 10^{-6}$. Note that the number of search sequences in the set $\mathcal{D}$ is typically a few more than the number of iterations of Algorithm \ref{al:minExpCost}, because $\mathcal{D}$ is initialized with $n$ search sequences, and Algorithm \ref{al:minExpCost} may add several extra search sequences to $\mathcal{D}$ in step \ref{step:solveandcheck} via Algorithm \ref{al:get_interior}.

\begin{table}[htb!]
\caption{The mean (95th percentile) of the number of search sequences required in $\mathcal{D}$ for convergence of $L$ and $U$ in Algorithm \ref{al:minExpCost}, for the \textit{varied} sample scheme.} \label{tab:sseqsconveps} 
\begin{center}
\begin{tabular}{|c|c c|c c|}
\hline
& \multicolumn{2}{|c|}{Acyclic}& \multicolumn{2}{|c|}{Cyclic}\\ 
\cline{2-5}
$n$ & $\epsilon=10^{-3}$ & $\epsilon=10^{-6}$ & $\epsilon=10^{-3}$ & $\epsilon=10^{-6}$ \\ 
\hline
2  & 5.15 (6) & 7.19 (9) & 5.03 (6)  & 5.70 (8)  \\
3  & 9.47 (12)	& 15.0 (20) & 8.96 (12) & 11.8 (17)\\
5  & 22.9 (28)	& 38.9 (50) & 20.8 (27) & 30.9 (43)\\
8  & 51.7 (62)	& 91.1 (116)&	44.0 (59) & 68.0 (98) \\
\hline
\end{tabular}
\end{center}
\end{table}

We next assess the quality of $\mathbf{p}_0$ as a heuristic for the hider.
Table \ref{tab:results} shows the decrease from $v^*$ to $v(\mathbf{p}_0)$ as a percentage of $v^*$ for the sample schemes in Table \ref{tab:dpstvar}, with $v^*$ either deduced to be equal to $v(\mathbf{p}_0)$ by Proposition \ref{prop:subgamesoln}, or otherwise computed by Algorithm \ref{al:minExpCost} with $\epsilon =  10^{-6}$.

As seen in Table \ref{tab:results}, $\mathbf{p}_0$ generally performs well as a hiding heuristic for a range of $n$ and, for smaller $n$, often achieves optimality. With the \textit{varied} sample scheme, for each $n$, in 95\% of acyclic games $\mathbf{p}_0$ is within 1.78\% of optimality, with this figure falling to 1.03\% for cyclic games. 
Therefore, if the hider cannot run Algorithm \ref{al:minExpCost} to estimate $\mathbf{p}^*$, the easily-calculated $\mathbf{p}_0$ performs well as a heuristic.

\begin{table}[htb!]
\small
\caption[caption]{The mean and percentiles of $v(\mathbf{p}_0)$ as percentage below optimum, and the percentage of search games in which $\mathbf{p}_0$ is optimal, for the sample schemes in Table \ref{tab:dpstvar}.
} \label{tab:results}
\begin{center}
\begin{tabular}{|c|c| c c c c|c c c c|}
\hline
 & & \multicolumn{4}{|c|}{Acyclic}& \multicolumn{4}{|c|}{Cyclic}\\ 
\cline{3-10} 
$n$ & Metric &Varied & Low & Medium & High &Varied & Low & Medium & High\\ 
\hline
2 & Mean    & 0.322 &  0.0733  & 0.0581  & 0.0357  & 0.163  &  0.0412  & 0.0352 &0.0195  \\
&75th Percentile & 0.493 & 0.119     & 0.0401    & 0     & 0.125  & 0.0486     & 0     & 0     \\
&95th Percentile & 1.43 &  0.291 & 0.363 & 0.213 & 0.991 &  0.223 & 0.240 &0.0681 \\
&99th Percentile & 2.00  & 0.366 & 0.649 & 0.881 & 1.69 & 0.313  & 0.563 & 0.616 \\
& \% $\mathbf{p}_0$ optimal    & 43.0 & 29.6 & 64.0 & 87.0  & 60.2  & 55.2 & 80.2 & 93.9 \\
\hline
3& Mean    & 0.537  &  0.0992   & 0.0524  & 0.0135  & 0.208 &   0.0492 &  0.0286  & 0.0076  \\
&75th Percentile      & 0.864 & 0.158  &   0.0444   & 0     & 0.266 & 0.0723     & 0     & 0     \\
&95th Percentile      & 1.72 &  0.31  & 0.301 & 0.0401 & 1.03&   0.212 &  0.200 & 0 \\
&99th Percentile      & 2.34  & 0.422 & 0.545 & 0.408 & 1.60 & 0.329 & 0.390 & 0.218 \\
& \% $\mathbf{p}_0$ optimal    & 21.4  & 12.7 & 55.7 & 91.7 & 45.4 & 39.5 & 77.2 &  96.3 \\
\hline
5& Mean    & 0.742 &  0.128  & 0.0441& 0.0012  & 0.273  &  0.0588& 0.0209 &0.0004  \\
&75th Percentile  & 1.10 & 0.192    & 0.0519     & 0     & 0.416 &0.0895       & 0     & 0     \\
&95th Percentile     & 1.77 &  0.319  & 0.211 & 0 & 1.01 & 0.200 & 0.135 &0 \\
&99th Percentile  & 2.35  & 0.399 & 0.407 & 0.0353 & 1.59 & 0.303 & 0.297 & 0 \\
& \% $\mathbf{p}_0$ optimal    & 7.06&  4.28 &44.4	&97.5 & 26.0  & 18.8 & 74.8 &99.2 \\
\hline
8  & Mean  & 0.882	&0.148	&0.0334	&0	&0.316&	0.0672	&0.0161&	0  \\
&75th Percentile & 1.20	& 0.204	&0.0440	&0	&0.470	&0.0994	&0.0064	&0     \\
&95th Percentile       & 1.78	&0.303	&0.147	&0	&0.953	&0.183	&0.0964	&0 \\
&99th Percentile       & 2.25	&0.373	&0.260	&0	&1.38	&0.259	&0.228	&0 \\

\hline
\end{tabular}
\end{center}
\end{table}

Table \ref{tab:results} shows that the optimality and performance of $\mathbf{p}_0$ depends strongly on $n$, the sample scheme and whether the search game is cyclic or acyclic. We first explain the patterns in sample scheme and search game type evident for each fixed $n$.

\paragraph{Patterns for fixed $n$}
We begin by making the following observation about $\mathbf{p}_0$ noted by \cite{GitRob1} for $n=2$. \cite{Norris1962} shows that if the hider was free to change boxes after every unsuccessful search, it is optimal for the hider to choose a new box according to $\mathbf{p}_0$, independent of previous hiding locations. Note that, if the hider plays $\mathbf{p}\equiv(p_1,\ldots,p_n)$ and the searcher first searches box $i$, then $p_iq_i/t_i$ is the detection probability per unit time of this first search, $i=1,\ldots, n$. Since $\mathbf{p}_0$ equates these terms and hence gives the searcher no preference of a box to search, the result in \cite{Norris1962} is not surprising.

Another way to interpret Norris' result is that it is optimal for the hider to keep the Gittins indices in \eqref{eq:simpleGI} equal throughout the search process. In our search game, the hider hides once at the start of the search, so it is impossible for the hider to maintain equality in \eqref{eq:simpleGI} after every unsuccessful search.
Intuitively, the best the hider can do is hide with probability $\mathbf{p}^*$ such that, when the searcher follows a Gittins search sequence against $\mathbf{p}^*$, the indices in \eqref{eq:simpleGI} are, on average, as close to being equal as possible throughout the search. However, the average should be weighted towards the start of the search, since the probability that the hider remains undetected decreases as time passes. Therefore, it is more important for the hider to achieve equality in \eqref{eq:simpleGI} earlier in the search rather than later, explaining why $\mathbf{p}_0$ is, in general, a good heuristic for the hider.


The preceding argument explains the following patterns in Table \ref{tab:results}. We see an improvement in performance of $\mathbf{p}_0$ in the \textit{high} sample scheme compared to the \textit{medium} sample scheme compared to the \textit{low} sample scheme, because the larger the detection probabilities, the sooner the hider is likely to be detected, and hence equality in \eqref{eq:simpleGI} near the start of the search takes even more importance. 

Further, recall that in a cyclic search game, where \eqref{eqn:cyclic} holds, if the hider plays $\mathbf{p}_0$ and the searcher uses any Gittins search sequence against $\mathbf{p}_0$, after $\sum_{i=1}^n x_i$ searches, equality in \eqref{eq:simpleGI} is reattained. 
Therefore, starting at $\mathbf{p}_0$, the Gittins indices tend to stay close together for cyclic games throughout the search process,
explaining why $\mathbf{p}_0$ performs better in cyclic than in acyclic search games.
This observation also explains why Algorithm \ref{al:minExpCost}, which starts with Gittins search sequences against $\mathbf{p}_0$, is seen to converge faster for cyclic games than acyclic games in Table \ref{tab:sseqsconveps}.


We also see an improvement in the performance of $\mathbf{p}_0$ in the \textit{medium} sample scheme, with its narrow range of detection probabilities, compared to the \textit{varied} sample scheme. To explain this phenomenon, we make the following connection to \cite{Clarkson:2020}, where the searcher knows the strategy of the hider, but has a choice between two search modes when searching any box.

In our search game, the hider chooses $\mathbf{p}$ to make the search last as long as possible, which involves balancing maximizing uncertainty about their location and forcing the searcher into boxes with ineffective search modes. \cite{Clarkson:2020} introduces two measures of the effectiveness of the search mode $(q_i,t_i)$ of a box $i$. The first, called the \emph{immediate benefit}, is measured by $q_i/t_i$. 
The larger the immediate benefit of box $i$, the greater the detection probability per unit time when box $i$ is searched. The second, called the \emph{future benefit}, is measured by
\begin{equation} \label{eqn:fbenefit}
\frac{-\log(1-q_i)}{t_i}.
\end{equation}
If $p_i=p_j$ and box $i$ has a larger future benefit than box $j$, then an unsuccessful search of box $i$ gains more information per unit time about the hider's location than an unsuccessful search of box $j$.
Whilst $\mathbf{p}_0$ takes the immediate benefit of the $n$ boxes' search modes into account by hiding in box $i$ with probability proportional to $t_i/q_i$, the future benefit is ignored by $\mathbf{p}_0$. 

In the game studied in \cite{Norris1962}, the hider may move between boxes after every unsuccessful search, so the game resets after every failed search.
Consequently, $\mathbf{p}_0$ is optimal since information gained by the searcher about the hider's location through an unsuccessful search is useless and hence the future benefit does not apply.
In our search game, however, the hider may not move between boxes, so gaining more information about the hider's location is useful, as it enables the searcher to make better box choices later in the search. Therefore, the hider should be dissuaded from hiding in boxes with a large future benefit, as the information-gain advantages of these boxes will benefit the searcher.

Since $\mathbf{p}_0$ does not take future benefit into account, the larger the variation in future benefit between the $n$ boxes, the worse $\mathbf{p}_0$ performs. With the \textit{varied} sample scheme, there is more opportunity for such variation, so $\mathbf{p}_0$ performs worse here than in the narrower \textit{medium} sample scheme.




\paragraph{Patterns as $n$ varies}
Table \ref{tab:results} shows, for the general \textit{varied} sample scheme, the performance of $\mathbf{p}_0$ degrades as $n$ increases, since the more boxes there are, the greater the uncertainty in the hider's location and hence, as demonstrated by \cite{Clarkson:2020}, the more valuable information about the hider's location becomes. Therefore, the future benefit, ignored by $\mathbf{p}_0$, takes more importance as $n$ grows, so the worse $\mathbf{p}_0$ performs.

%
%

However, Table \ref{tab:results} also shows that the change in performance of $\mathbf{p}_0$ with $n$ is strongly affected by the underlying sample scheme.  
As previously noted, the smaller the variation in future benefit between the $n$ boxes, the better $\mathbf{p}_0$ will perform.
The narrower the range of detection probabilities in the sample scheme, the quicker the variation in future benefit decreases as we add more boxes. Therefore, whilst the degradation in $\mathbf{p}_0$ as $n$ increases due to more importance on future benefit still dominates for the \textit{varied} sample scheme, it is nullified by this decrease in future benefit variation for the narrower \textit{medium} sample scheme, so much so that in Table \ref{tab:results} the performance of $\mathbf{p}_0$ slightly improves with $n$ for the \textit{medium} sample scheme.

Further, the smaller the detection probabilities, the longer the search is expected to last, so the more important future box choices become. Hence, the smaller the detection probabilities, the greater the importance of the future benefit and the worse $\mathbf{p}_0$ will perform. Therefore, for the \textit{low} sample scheme, we see the sharpest decline in the performance of $\mathbf{p}_0$ as $n$ increases. Due to large detection probabilities and a narrow sample scheme, the performance of $\mathbf{p}_0$ improves strongly with $n$ in the \textit{high} sample scheme.

\subsection{Future Benefit for Two-Box Problems}
In this section, for $n=2$, we examine the effect of the difference between the future benefit of the two boxes on the difference between $\mathbf{p}_0\equiv (p_0,1-p_0)$ and the optimal hiding strategy $\mathbf{p}^*\equiv (p^*,1-p^*)$. 

\cite{GitRob1} studied two-box search games with $q_1 < q_2$ and unit search times, noting that whenever $\mathbf{p}_0$ was suboptimal, $p^*$ was greater than $p_0$, but found no reason for this observation. We believe this phenomenon is explained by future benefit. Since $q_1<q_2$ and $t_1=t_2=1$, the future benefit in \eqref{eqn:fbenefit} at any $\mathbf{p}$ is greater for box 2 than box 1. Whilst $\mathbf{p}_0$ considers immediate benefit, it ignores future benefit, explaining why the hider, who wants the searcher to spend more time in boxes with inefficient search modes, may prefer to hide in box 1 with a probability greater than $p_0$.

\begin{figure}[h!]
\begin{center}
\includegraphics[scale=0.58]{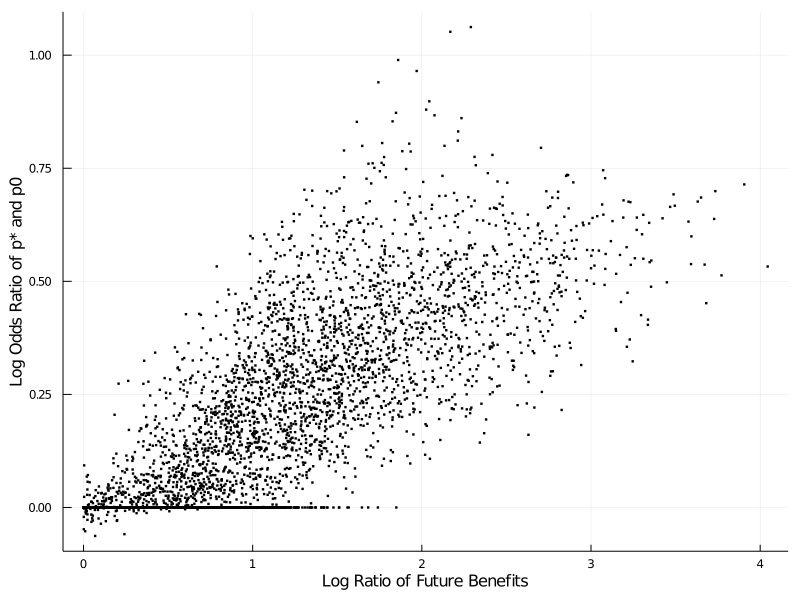}
\caption{The relationship between the terms in \eqref{eqn:scatterdata}: the log of the ratio of the future benefits of the two boxes is on the horizontal axis and the log odds ratio of $p^*$ and $p_0$ is on the vertical axis.}
\label{fig:log_future_benefit}
\end{center}
\end{figure}

To demonstrate this effect, we conduct an additional numerical study with $n=2$. The two boxes are drawn using \eqref{eqn:acyclicdraw} to generate an acyclic search game, then relabelled so box 1 has the  lower future benefit in \eqref{eqn:fbenefit}. We then calculate
\begin{equation} \label{eqn:scatterdata}
\log\left(\frac{\log(1-q_2)t_1}{\log(1-q_1)t_2}\right) \quad \text{and} \quad \log\left(\frac{p^*(1-p_0)}{p_0(1-p^*)}\right),
\end{equation}
the former the log of the relative increase in future benefit from box 1 to box 2, and the latter the log odds ratio of $p^*$ and $p_0$, the logarithm used to increase the visibility of a discrepancy between values $p^*$ and $p_0$ either close to 0 or 1. We repeat the preceding $5,000$ times. Figure \ref{fig:log_future_benefit} shows the relationship between the two values calculated in \eqref{eqn:scatterdata}, the former on the horizontal axis and the latter on the vertical axis. 

Figure \ref{fig:log_future_benefit} shows that almost all log odds ratios are positive (only 0.96\% are negative), indicating that $p^* \geq p_0$ in more than 99\% of problems where box 1 has the smaller future benefit. Of those 40\% of problems with zero vertical coordinate, corresponding to $p^* = p_0$, the mean log ratio of future benefit was 0.45, showing only a small increase in future benefit from box 1 to box 2. On the other hand, the same mean was 1.4 for those problems with strictly positive vertical coordinate, corresponding to $p^* > p_0$, showing a larger increase in future benefit from box 1 to box 2.
Further, Figure \ref{fig:log_future_benefit} demonstrates a clear positive relationship between the terms in \eqref{eqn:scatterdata}, showing the greater the increase in future benefit from box 1 to box 2, the farther $p^*$ is above $p_0$.


In addition, \cite{Ruckle:1991} solves a two-box game with $t_1=t_2=q_2=1$ and a sole parameter $q_1 \equiv q \in (0,1)$. In this problem
$$p_0 = \frac{1/q}{1/q+1}.$$

\cite{Ruckle:1991} shows the hider optimally hides in box 1 with probability
\begin{equation} \label{eqn:ruckleopt}
p^* \equiv \frac{1/q}{1/q+(1-q)^{h-1}} \quad \text{where} \quad h \equiv \floor{\bar{h}} \quad \text{and} \quad \bar{h}=1/q+(1-q)^{\bar{h}-1}.
\end{equation}
We analyse this result of \cite{Ruckle:1991} to conclude the following.
The optimal hiding strategy in \eqref{eqn:ruckleopt} leads to a tie between the Gittins indices of the boxes for the $h$th search, and $p_0$ is optimal for the hider if and only if $h=1$. As $q$ decreases, $h$ increases, so $p^*$ increases; see Table \ref{tab:Ruckle}.

\begin{table}[h!]
\caption{For the two-box game with $t_1=t_2=q_2=1$ and $q_1 \equiv q \in (0,1)$, the value of $h$ in \eqref{eqn:ruckleopt} by the value of $q$.} \label{tab:Ruckle}
\begin{center}  
\begin{tabular}{|c| c|} 
 \hline
Value of $h$ & Range of $q$ \\
\hline
1 & $[0.618,1]$ \\
2    & $[0.382,0.618]$ \\
3    & $[0.276,0.382]$ \\
\hline
\end{tabular}
\end{center}
\end{table}

We offer the following explanation. Recall hiding with $p_0$ ignores future benefit. For any $q \in (0,1)$, the future benefit in \eqref{eqn:fbenefit} is greater for box 2 than for box 1, with the size of the difference growing as $q$ decreases. 
Therefore, for larger $q$, the hider optimally hides in box 1 with probability $p_0$, since the difference in future benefit between the two boxes is small. As $q$ decreases, the advantage in future benefit of box 2 over box 1 grows. Wishing the searcher to spend more time in boxes with poor search modes, the hider optimally hides in box 1 with a probability $p^*$ increasingly larger than $p_0$.

\cite{Ruckle:1991} also finds that an optimal search strategy is a mixture of the two search sequences which make their only search of box 2 on the $h$th (resp. $(h+1)$st) search. Since any Gittins search sequence against $p^*$ encounters its only tie on its $h$th search, Ruckle's optimal search strategy is a mixture of the two elements of $\widehat{\mathcal{C}}_{p^*}$, so satisfies Theorem \ref{thm:OSSperm}.


\section{Conclusion} \label{sec:conclusion}

This paper develops very significantly the existing literature on a search game in discrete boxes where the searcher may overlook a well-concealed hider.
There are theoretical links to the problem where the hider is replaced by an inanimate object hidden randomly by Nature. In this problem, the searcher optimally exploits boxes most attractive to them. An intelligent hider will all but take away the notion of one box being more attractive than another, with the searcher's focus now on randomizing their strategy to guard against being taken advantage of by the hider.

Since a pure strategy for the searcher is an indefinite list of boxes to search until the hider is found, the search game is semi-finite and hence difficult to analyse. As a result, most work in the current literature is limited to two boxes or boxes searched in unit time. 
Using novel proof techniques, 
we develop a comprehensive theory for the fully-general search game by extending much of the existing work and uncovering new properties along the way.

By making an adjustment to the set of search strategies, we provide a rigorous proof that an optimal search strategy exists, extending a result of \cite{Bram1963}. We next develop properties of an optimal search strategy, and, extending a two-box result of \cite{Gittins:1989}, we show that the searcher can construct an optimal strategy by randomly choosing between some $n$ of $n!$ known, simple search strategies.
Based on these properties, we present a novel practical procedure to test if any hiding strategy is optimal, which we use in a numerical study to investigate the frequency of the optimality of a particular hiding strategy that gives the searcher no preference over any box at the beginning of the search. We interpret the patterns in our results to obtain valuable insight into optimal hiding strategies, which will aid the construction of effective search strategies.

Further work may include a search game on a network structure rather than in discrete boxes. Such an extension is relevant if the geography of the search space prevents the searcher from moving quickly between any pair of hiding locations, for example, a structure of roads. Search games on networks are well studied in the literature, but less so with a chance of overlook. 

\section*{Acknowledgements}
We are grateful for the support of the EPSRC funded EP/L015692/1 STOR-i Centre for Doctoral Training.
The authors thank Dashi Singham and Steve Alpern for many helpful discussions and suggestions.

\clearpage

\begin{appendices}
\section{Proof of Lemma \ref{lemma:altminimaxbandhider}} \label{append:closedlemma}
By Theorem 2.4.2 of \cite{Blackwell1954}, Lemma \ref{lemma:altminimaxbandhider} holds if $\mathcal{S}(\epsilon)$ is closed.
Write
$$\bar{\mathcal{S}}(\epsilon) \equiv \mathcal{S}(\epsilon) \setminus \{(V_1(\zeta_i(\epsilon)),\ldots, V_n(\zeta_i(\epsilon))), \; i=1,\ldots, n\}.$$
Since they differ by a finite subset of $\mathbb{R}^n$, $\mathcal{S}(\epsilon)$ is closed if and only if $\bar{\mathcal{S}}(\epsilon)$ is closed. Therefore, the proof will be completed by showing that $\bar{\mathcal{S}}(\epsilon)$ is closed. 


Throughout the proof, write $V(\xi)\equiv (V_1(\xi), \ldots , V_n(\xi))$; therefore, any element of $\bar{\mathcal{S}}(\epsilon)$ takes the form $V(\xi)$ where $\xi$ is a 
Gittins search sequence against some $\mathbf{p} \in \mathcal{P}(\epsilon)$.

By Definition \ref{def:GIP}, the next box searched by any Gittins search sequence against a hiding strategy $\mathbf{p}$ must satisfy \eqref{eq:simpleGI}. If, at some point whilst following a Gittins search sequence against $\mathbf{p}$, multiple boxes satisfy \eqref{eq:simpleGI}, we say the searcher has encountered a \emph{tie} and $\mathbf{p}$ is a \emph{tie point}. Note that an equivalent definition of a tie point $\mathbf{p}$ is $|\mathcal{C}_{\mathbf{p}}|>1$. If $|\mathcal{C}_{\mathbf{p}}|=1$, we say $\mathbf{p}$ is a \emph{non-tie point}. If $\mathbf{p}$ is a non-tie point, then there is a unique Gittins search sequence against $\mathbf{p}$, whereas, for a tie point $\mathbf{p}$, a specific Gittins search sequence against $\mathbf{p}$ is determined by how we break ties between boxes.
 
Let the set of rules for breaking ties be $\mathcal{R}$. We can think of $\mathcal{R}$ as a set of infinite sequences whose elements are permutations of $\{1,\ldots , n\}$. The $j$th element of $\mathbf{r} \in \mathcal{R}$ is the preference ordering with which the $j$th encountered tie should be broken. 
For example, suppose $n=5$, and the $j$th tie encountered following a Gittins search sequence against $\mathbf{p}$ involves boxes 2, 3 and 5. Suppose the $j$th element of $\mathbf{r}$ is $54231$. Then, under rule $\mathbf{r}$, tie $j$ is split by searching boxes 5, 2 and 3 in that order. Note that changing the $j$th element of $\mathbf{r}$ to $45123$ does not affect the Gittins search sequence generated, demonstrating that multiple rules can generate the same Gittins search sequence.


Further, note how $\mathcal{R}$ can be identified with the interval $[0,1]$ in the following way.
Any term in any $\mathbf{r} \in \mathcal{R}$ is one of the $n!$ elements of $S_n$, where $S_n$ is the set of permutations of $\left\{ 1,\ldots, n\right\}$. Number the elements of $S_n$ from $0$
to $n!-1$, and rewrite $\mathbf{r}$ as $x_1x_2\ldots x_j\ldots $, where $x_j$ is the number (from $0$ to $n!-1$) representing the $j$th element of $\mathbf{r}$. We now associate with $\mathbf{r}$ the number in $[0,1]$ given by 
\[
\phi (\mathbf{r}) = \sum_{j=1}^\infty \frac{x_j}{(n!)^j}.
\]
The mapping $\phi :\mathcal{R} \rightarrow [0,1] $ is a bijection. Therefore, by a convergent subsequence $\left\{\mathbf{r}_a : a \in \mathbb{Z}^+\right\}$ in $\mathcal{R}$, we mean a sequence for which $\left\{ \phi ( \mathbf{r}_{a}): a \in \mathbb{Z}^+ \right\}$ converges in $[0,1]$, where $\mathbb{Z}^+$ is the set of strictly positive integers. However, for the remainder of the proof, we shall continue to interpret $\mathcal{R}$ as a set of infinite sequences with elements in $S_n$.

%
%

Write
$$\mathcal{P}\equiv \left\{(p_1, \ldots , p_n): p_i \geq 0, \; \sum_{i=1}^n p_i=1\right\}$$
for the space of mixed hiding strategies.
Write $f$ for the function from $\mathcal{R} \times \mathcal{P} \rightarrow \bar{\mathcal{S}}(\epsilon)$ satisfying $f(\mathbf{r},\mathbf{p})=V(\xi(\mathbf{r},\mathbf{p}))$, where $\xi(\mathbf{r},\mathbf{p})$ is the Gittins search sequence against $\mathbf{p}$ that breaks ties using rule $\mathbf{r}$. In other words, $f$ maps a hiding strategy and tie-breaking rule to the vector of conditional expected times to detection of the corresponding Gittins search sequence.


Before showing that $\bar{\mathcal{S}}(\epsilon)$ is closed, we first show the closure of a smaller set, concerning only Gittins search sequences against a fixed hiding strategy with $p_i>0$ for $i=1,\ldots , n$.


\begin{lemma} \label{lemma:fixedpclosed}
For any $\mathbf{p} \in \mathcal{P}$ with $p_i>0$ for $i=1,\ldots , n$, the set
\begin{equation} \label{eqn:S(p)defn}
\mathcal{S}_{\mathbf{p}}\equiv \{V(\xi)
: \xi \in \mathcal{C}_{\mathbf{p}}\}
\end{equation}
is closed. 
\end{lemma}
\begin{proof}
Since $p_i>0$ for $i=1,\ldots , n$, we have $\mathcal{S}_{\mathbf{p}} \subset \mathbb{R}^n$.
There are two cases. First, if $|\mathcal{S}_{\mathbf{p}}|$ is finite, then $\mathcal{S}_{\mathbf{p}}$ is a finite collection of points in $\mathbb{R}^n$, so $\mathcal{S}_{\mathbf{p}}$ is closed.
%
Second, if $|\mathcal{S}_{\mathbf{p}}|$ is infinite, then $\mathbf{p}$ must be a tie point.
To show that $\mathcal{S}_{\mathbf{p}}$ is closed, we need to show that any convergent sequence $\{\mathbf{s}_a: a \in \mathbb{Z}^+\}$ in $\mathcal{S}_{\mathbf{p}}$ must have its limit  $\mathbf{s}_0 \equiv \lim_{a \rightarrow \infty} \mathbf{s}_a$ also in $\mathcal{S}_{\mathbf{p}}$.


Since any element of $\mathcal{S}_{\mathbf{p}}$ corresponds to some rule in $\mathcal{R}$, the image of the function $f$ with second argument fixed at $\mathbf{p}$ is equal to $\mathcal{S}_{\mathbf{p}}$; therefore, 
for each $a \in \mathbb{Z}^+$, we can choose $\mathbf{r}_a \in \mathcal{R}$ such that $\mathbf{s}_a=f(\mathbf{r}_a, \mathbf{p})$.
Consider the sequence $\{\mathbf{r}_a : a \in \mathbb{Z}^+\}$ and, by identifying $\mathcal{R}$ with the interval $[0,1]$, choose a convergent subsequence $\{\mathbf{r}_{h(a)}:a \in \mathbb{Z}^+\}$.

Identifying $\mathcal{R}$ with the interval $[0,1]$  shows $\mathcal{R}$ is closed;
therefore, $\mathbf{r}_0 \equiv \lim_{a \rightarrow \infty} \mathbf{r}_{h(a)} \in \mathcal{R}$, so $f(\mathbf{r}_0, \mathbf{p}) \in \mathcal{S}_{\mathbf{p}}$.
Any infinite subsequence of the convergent sequence $\{\mathbf{s}_a\}$ must converge to the same limit as $\{\mathbf{s}_a\}$; therefore, we have $\mathbf{s}_0 = \lim_{a \rightarrow \infty} \mathbf{s}_{h(a)}$. To complete the proof, we show that $\mathbf{s}_0 = f(\mathbf{r}_0, \mathbf{p})$, so $\mathbf{s}_0 \in \mathcal{S}_{\mathbf{p}}$.

To ease notation for the remainder of the proof, since $\mathbf{p}$ is fixed, we drop the second argument from $f$ and $\xi$.
Therefore, we have $\mathbf{s}_{h(a)}=f(\mathbf{r}_{h(a)})=V(\xi(\mathbf{r}_{h(a)}))$, $f(\mathbf{r}_0)=V(\xi(\mathbf{r}_0))$,
and our aim to show that $\mathbf{s}_0 = f(\mathbf{r}_0)$ is equivalent to showing that $\lim_{a \rightarrow \infty} V(\xi(\mathbf{r}_{h(a)}))=V(\xi(\mathbf{r}_0))$.

Since $\mathbf{r}_0 \equiv \lim_{a \rightarrow \infty} \mathbf{r}_{h(a)}$, for each $j \in \mathbb{Z}^+$, there must exist a smallest element of $\mathbb{Z}^+$, say $a_j$, such that the first $j$ elements of $\mathbf{r}_{h(a)}$ are equal to the first $j$ elements of $\mathbf{r}_{0}$ for all $a \geq a_j$. Further, the $\{a_j : j \in \mathbb{Z}^+\}$ must form an increasing sequence. Therefore, for any $a \geq a_j$, both $\xi(\mathbf{r}_{h(a)})$ and $\xi(\mathbf{r}_0)$ break the first $j$ ties encountered by any Gittins search sequence against $\mathbf{p}$ in the same manner, so, as $j$ increases, the first time when $\xi(\mathbf{r}_{h(a_j)})$ and $\xi(\mathbf{r}_0)$ differ becomes increasingly later and later into the search. Hence, no matter where the hider is hidden, the effect on the expected time to detection of this difference decreases to 0; in other words, for $i = 1,\ldots , n$, we have $\lim_{j \rightarrow \infty} V_i(\xi(\mathbf{r}_{h(a_j)})) \rightarrow V_i(\xi(\mathbf{r}_0))$, so $\lim_{j \rightarrow \infty}V(\xi(\mathbf{r}_{h(a_j)})) = V(\xi(\mathbf{r}_0))$. Since the $\{a_j : j \in \mathbb{Z}^+\}$ form an increasing sequence in $\mathbb{Z}^+$, we have $\lim_{j \rightarrow \infty}V(\xi(\mathbf{r}_{h(a_j)})) = \lim_{a \rightarrow \infty}V(\xi(\mathbf{r}_{h(a)}))$, completing the proof.
\end{proof}



Next, via two lemmas, we investigate the continuity of the function $f(\mathbf{r},\mathbf{p})$ in its second argument at different $\mathbf{p}$ for any fixed $\mathbf{r} \in \mathcal{R}$. The first lemma deals with the simpler case of continuity at non-tie points in $\mathcal{P}(\epsilon)$. 

\begin{lemma} \label{lemma:frcont}
If $\mathbf{p} \in \mathcal{P}(\epsilon)$ is a non-tie point, then, for any fixed first argument $\mathbf{r} \in \mathcal{R}$, $f$ is continuous in its second argument at $\mathbf{p}$.
\end{lemma}
\begin{proof}
Write $\xi$ for the unique Gittins search sequence against $\mathbf{p}$. For $k \in \mathbb{Z}^+$, write $\mathcal{P}_k \subset \mathcal{P}$ for the set of mixed hiding strategies $\mathbf{x}$ for which every Gittins search sequence against $\mathbf{x}$ is identical to $\xi$ for the first $k$ searches. Clearly, for any $k \in \mathbb{Z}^+$, we have $\mathcal{P}_{k+1} \subseteq \mathcal{P}_k$ and $\mathbf{p} \in \mathcal{P}_k$. 

For any $\delta >0$, write $B(\mathbf{p}, \delta)$ for the open ball with radius $\delta$ centred at $\mathbf{p}$. Since $\mathbf{p}$ is a non-tie point in $\mathcal{P}(\epsilon)$, for any $k \in \mathbb{Z}^+$, it is possible, in any direction, to move a small-enough (Euclidean) distance in $\mathcal{P}$ away from $\mathbf{p}$ and not disrupt the order of the Gittins indices that generate the first $k$ searches of $\xi$. 
Therefore,
\begin{equation} \label{eqn:delta_kdef}
\delta_k \equiv 0.5 \times \sup \left\{\delta : B(\mathbf{p}, \delta) \subseteq \mathcal{P}_k \right\} > 0,
\end{equation}
with $\delta_{k} \geq \delta_{k+1}$, and 0.5 chosen arbitrarily in $(0,1)$ to ensure that $B(\mathbf{p}, \delta_k) \subseteq \mathcal{P}_k$ for all $k \in \mathbb{Z}^+$.

Write $\delta^* \equiv \lim_{k \rightarrow \infty} \delta_k$.
There are two cases. First suppose that $\delta^*>0$. 
If $\mathbf{x} \in B(\mathbf{p},\delta^*)$, then $\mathbf{x} \in \mathcal{P}_k$ for all $k \in \mathbb{Z}^+$, so any Gittins search sequence against $\mathbf{x}$ is identical to $\xi$. It follows that 
$\xi$, the unique Gittins search sequence against $\mathbf{p}$, is also the unique Gittins search sequence against $\mathbf{x}$.
Hence, $f$ is constant on $\mathcal{R} \times
B(\mathbf{p},\delta^*)$, so $f$ is continuous in its second argument at $\mathbf{p}$.

Second, suppose that $\delta^*=0$. Consider a sequence $\{\mathbf{x}_a\}$ in $\mathcal{P}$ with $\lim_{a \rightarrow \infty} \mathbf{x}_a = \mathbf{p}$. To show $f$ is continuous in its second argument at $\mathbf{p}$, we show that
\begin{equation} \label{eqn:toshowcont}
\lim_{a \rightarrow \infty} f (\mathbf{r},\mathbf{x}_a) = f(\mathbf{r},\mathbf{p})
\end{equation}
for any fixed $\mathbf{r} \in \mathcal{R}$.

For any $k \in \mathbb{Z}^+$, since $\delta_k>0$, there must exist a smallest number $g(k) \in \mathbb{Z}^+$ such that every term in the sequence $\{\mathbf{x}_a\}$ after $\mathbf{x}_{g(k)}$ belongs to the ball $B(\mathbf{p},\delta_k)$. Formally, for any $k \in \mathbb{Z}^+$, write
\begin{equation} \label{eqn:g(k)def}
g(k) \equiv \min \{A:\mathbf{x}_a \in B(\mathbf{p},\delta_k), \;a \geq A\}.
\end{equation}
Since $\delta_k \geq \delta_{k+1}$, we have $B(\mathbf{p},\delta_{k+1}) \subseteq B(\mathbf{p},\delta_{k})$ and hence $g(k) \leq g(k+1)$, so the sequence $\{g(k): k \in \mathbb{Z}^+\}$ increases weakly. 


Consider the sequence $\{\mathbf{x}_{g(k)} : k \in \mathbb{Z}^+\}$. We have $\lim_{a\rightarrow \infty} \mathbf{x}_a=\mathbf{p}$ by assumption; our next aim is to show that $\lim_{k\rightarrow \infty} \mathbf{x}_{g(k)} = \mathbf{p}$ also. 
To do this, we show that, for any $\epsilon>0$, we can choose $K$ such that $\mathbf{x}_{g(k)} \in B(\mathbf{p},\epsilon)$ for all $k \geq K$. Choose $\epsilon>0$. Since $\lim_{k \rightarrow \infty} \delta_k = 0$, there exists $K$ such that $\delta_K < \epsilon$. 
By the definition of $g$ in \eqref{eqn:g(k)def}, we have $\mathbf{x}_a \in B(\mathbf{p}, \delta_K)$ for all $a \geq g(K)$. Since $g$ is increasing, we have $\mathbf{x}_{g(k)} \in B(\mathbf{p}, \delta_K) \subset B(\mathbf{p}, \epsilon)$ for all $k \geq K$, showing that $\{\mathbf{x}_{g(k)}\}$ has limit $\mathbf{p}$.

By the definitions in \eqref{eqn:delta_kdef} and \eqref{eqn:g(k)def}, we have $\mathbf{x}_{g(k)} \in B(\mathbf{p}, \delta_k) \subseteq \mathcal{P}_{k}$. Recall $\xi$ as the unique Gittins search sequence against $\mathbf{p}$, and, for $b \in \mathbb{Z}^+$, write $\xi_{b}$ for an arbitrary Gittins search sequence against $\mathbf{x}_b$. 
Since $\mathbf{x}_{g(k)} \in \mathcal{P}_{k}$, as $k$ increases, the first time when $\xi_{g(k)}$ and $\xi$ may differ becomes increasingly later and later into the search. Hence, no matter where the hider is hidden, the effect on the expected time to detection of this difference decreases to 0; in other words, for $i = 1,\ldots , n$, $V_i(\xi_{g(k)}) \rightarrow V_i(\xi)$ as $k \rightarrow \infty$, $i = 1,\ldots , n$, so
$$\lim_{k \rightarrow \infty} f (\mathbf{r},\mathbf{x}_{g(k)}) = f(\mathbf{r},\mathbf{p})$$
for any $\mathbf{r} \in \mathcal{R}$.
Since $\lim_{k\rightarrow \infty} \mathbf{x}_{g(k)}=\lim_{a\rightarrow \infty} \mathbf{x}_a=\mathbf{p}$, then \eqref{eqn:toshowcont} follows, completing the proof.
\end{proof}

Now we consider the continuity of $f$ in its second argument at tie points in $\mathcal{P}(\epsilon)$, the more challenging case. Informally, if $\mathbf{p} \in \mathcal{P}(\epsilon)$ is a tie point, then $f$ is only continuous in its second argument at $\mathbf{p}$ for certain fixed first arguments $\mathbf{r} \in \mathcal{R}$, and only approaching $\mathbf{p}$ via certain paths in $\mathcal{P}$.
 


To state the continuity conditions precisely, we first need a few definitions. 
Let $\mathbf{p} \equiv (p_1, \ldots , p_n)$ be a tie point in $\mathcal{P}(\epsilon)$. Recall $S_n$ as the set of permutations of $\{1,\ldots, n\}$, and let $\Sigma \subseteq S_n$.
For $\sigma \in S_n$, write $\sigma(i)$ for the number in the $i$th position of $\sigma$ and write
\begin{equation} \label{eqn:Psigmaset}
\mathcal{P}(\mathbf{p}, \Sigma) \equiv \left\{\mathbf{x} \equiv(x_1,\ldots,x_n) \in \mathcal{P}: x_{\sigma(i)}/p_{\sigma(i)} \geq x_{\sigma(i+1)}/p_{\sigma(i+1)}, \; i=1,2,\ldots, n-1, \; \sigma \in \Sigma \right\}.
\end{equation}


The set $\mathcal{P}(\mathbf{p}, \Sigma)$ may be interpreted as follows. Suppose $\mathbf{x} \in \mathcal{P}(\mathbf{p},\Sigma)$. Then, for any $\sigma \in \Sigma$, if any tie encountered in a Gittins search sequence against $\mathbf{p}$ is broken using $\sigma$, the order that the tied boxes are searched remains the same if we replace $\mathbf{p}$ with $\mathbf{x}$ when calculating the Gittins indices in \eqref{eq:simpleGI} at this tie, and still use $\sigma$ to break any remaining ties. Clearly $\mathbf{p} \in \mathcal{P}(\mathbf{p}, \Sigma)$ for any subset $\Sigma$ of $S_n$.


Further, note that if $\mathbf{x},\mathbf{y} \in \mathcal{P}(\mathbf{p}, \Sigma)$, then for any $\lambda \in [0,1]$, we must have $\lambda \mathbf{x} + (1-\lambda) \mathbf{y} \in \mathcal{P}(\mathbf{p}, \Sigma)$. Therefore, $\mathcal{P}(\mathbf{p}, \Sigma)$ is a convex set containing $\mathbf{p}$ for any $\Sigma \subset S_n$.

Informally, the following lemma says that, if its first argument is fixed to be some $\mathbf{r}\in \mathcal{R}$ containing only elements of $\Sigma \subset S_n$, then $f$ is continuous in its second argument at $\mathbf{p}$ approaching from any path in $\mathcal{P}(\mathbf{p}, \Sigma)$.

\begin{lemma} \label{lem:frsemicont}
Suppose $\mathbf{p} \in \mathcal{P}(\epsilon)$ is a tie point and $\Sigma \subset S_n$.
Let $\{\mathbf{x}_a : a \in \mathbb{Z}^+\}$ be a sequence in $\mathcal{P}(\mathbf{p}, \Sigma)$ with $\lim_{a \rightarrow \infty} \mathbf{x}_a = \mathbf{p}$.
Then, for any $\mathbf{r} \in \mathcal{R}$ whose elements all belong to $\Sigma$, we have $\lim_{a \rightarrow \infty} f (\mathbf{r},\mathbf{x}_a) = f(\mathbf{r},\mathbf{p})$.
\end{lemma}

\begin{proof}
%
First, note that if $\mathcal{P}(\mathbf{p},\Sigma)=\{\mathbf{p}\}$, then any sequence in $\mathcal{P}(\mathbf{p}, \Sigma)$ is constant, and the result is trivially true. The rest of the argument, which is similar to the proof of Lemma \ref{lemma:frcont}, deals with the case where $\mathcal{P}(\mathbf{p},\Sigma)$ contains elements in addition to $\mathbf{p}$. Let $\mathbf{r} \in \mathcal{R}$ contain only elements from $\Sigma \subset S_n$. 
For $k \in \mathbb{Z}^+$, let $\mathcal{P}_{k,\mathbf{r}} \subset \mathcal{P}$ contain precisely those mixed hiding strategies $\mathbf{x}$ for which the Gittins search sequence against $\mathbf{x}$ under rule $\mathbf{r}$ is identical to $\xi(\mathbf{r},\mathbf{p})$ (the Gittins search sequence against $\mathbf{p}$ under rule $\mathbf{r}$) for the first $k$ searches. Clearly, for any $k \in \mathbb{Z}^+$, we have $\mathcal{P}_{k+1, \mathbf{r}} \subseteq \mathcal{P}_{k, \mathbf{r}}$ and $\mathbf{p} \in \mathcal{P}_{k, \mathbf{r}}$.

For any $\delta >0$, write $B(\mathbf{p}, \delta)$ for the open ball with radius $\delta$ centred at $\mathbf{p}$. 
Note that any two points in $\mathcal{P}$ must be within Euclidean distance $\sqrt{n}$ of eachother. Therefore, for any $\mathbf{p} \in \mathcal{P}$, we must have $B(\mathbf{p}, \sqrt{n}) = \mathcal{P}$. For $\delta \in [0,\sqrt{n}]$, write $\mathcal{P}(\mathbf{p}, \delta, \Sigma) \equiv B(\mathbf{p}, \delta) \cap \mathcal{P}(\mathbf{p}, \Sigma)$. 
In other words, $\mathcal{P}(\mathbf{p}, \delta, \Sigma)$ is the subset of mixed hiding strategies in $\mathcal{P}(\mathbf{p}, \Sigma)$ strictly less than (Euclidean) distance $\delta$ from $\mathbf{p}$.


Write
\begin{equation} \label{eqn:delta_k,r}
\delta_{k,\mathbf{r}} \equiv 0.5 \times \sup \left\{\delta : \mathcal{P}(\mathbf{p}, \delta, \Sigma) \subseteq \mathcal{P}_{k, \mathbf{r}} \right\},
\end{equation}
with 0.5 arbitrarily chosen in $(0,1)$ to ensure that $\mathcal{P}(\mathbf{p}, \delta_{k,\mathbf{r}}, \Sigma) \subseteq \mathcal{P}_{k, \mathbf{r}}$ for all $k \in \mathbb{Z}^+$.
Note that $\delta_{k,\mathbf{r}} \geq 0$ for all $k \in \mathbb{Z}^+$ since $\mathcal{P}(\mathbf{p}, 0, \Sigma) = \{ \mathbf{p}\} \in \mathcal{P}_{k, \mathbf{r}}$. 
The aim of the following is to show that $\delta_{k,\mathbf{r}} > 0$ for all $k \in \mathbb{Z}^+$.

Let $k \in \mathbb{Z}^+$. We examine two cases. First, suppose that, in the first $k$ searches of $\xi(\mathbf{r},\mathbf{p})$, no ties are encountered. Then, it is possible, in any direction, to move a small enough (Euclidean) distance in $\mathcal{P}$ away from $\mathbf{p}$ and not disrupt the order of the Gittins indices in \eqref{eq:simpleGI} that generate the first $k$ searches of $\xi(\mathbf{r},\mathbf{p})$. 
Therefore, we may choose $\delta>0$ such that $B(\mathbf{p},\delta) \subset \mathcal{P}_{k, \mathbf{r}}$. It follows that $\mathcal{P}(\mathbf{p},\delta, \Sigma) \subset \mathcal{P}_{k,\mathbf{r}}$, and hence that $\delta_{k,\mathbf{r}}\geq \delta/2>0$. 

Second, suppose that, in the first $k$ searches of $\xi(\mathbf{r},\mathbf{p})$, we do encounter ties between boxes. Suppose such a tie involves $b$ boxes. By \eqref{eq:simpleGI}, whilst the order of the next $b$ boxes searched may 
depend on the tie-breaking rule, the set of $b$ boxes searched will not. Therefore, after the tie has been broken, the Gittins indices in \eqref{eq:simpleGI} will be the same no matter how the tie was broken.
Hence, it is possible, in any direction, to move a small enough (Euclidean) distance away in $\mathcal{P}$ from $\mathbf{p}$ and not disrupt the order of the Gittins indices that generate the first $k$ searches of $\xi(\mathbf{r},\mathbf{p})$ at any point where there is not a tie between boxes. It follows that we may choose $\delta>0$ such that, for any $\mathbf{x} \in B(\mathbf{p}, \delta)$, any Gittins search sequence against $\mathbf{x}$ differs only in the first $k$ searches to $\xi(\mathbf{r},\mathbf{p})$ for those searches where $\xi(\mathbf{r},\mathbf{p})$ is in the process of breaking a tie.
Now suppose additionally that $\mathbf{x} \in \mathcal{P}(\mathbf{p}, \Sigma)$, so $\mathbf{x} \in \mathcal{P}(\mathbf{p}, \delta, \Sigma)$. 
Since $\mathbf{x} \in \mathcal{P}(\mathbf{p}, \Sigma)$, 
when a tie is reached by $\xi(\mathbf{r},\mathbf{p})$, if we instead were following a Gittins search sequence against $\mathbf{x}$, the Gittins indices of any boxes involved in the tie will either still be tied, or lie in the ordering determined by $\sigma$ for all $\sigma \in \Sigma$.
Therefore, since $\mathbf{r}$ contains only elements of $\Sigma$, the Gittins search sequence against $\mathbf{x}$ that breaks ties using $\mathbf{r}$ will break the tie using the same preference ordering as $\xi(\mathbf{r},\mathbf{p})$, so will be identical to $\xi(\mathbf{r},\mathbf{p})$ for the first $k$ searches. In other words, $\mathcal{P}(\mathbf{p}, \delta, \Sigma) \subset \mathcal{P}_{k,\mathbf{r}}$, and hence $\delta_{k,\mathbf{r}} \geq \delta/2>0$. 


Now we have shown $\delta_{k,\mathbf{r}} > 0$ for all $k \in \mathbb{Z}^+$, we are in a position to finish the proof in a similar style to Lemma \ref{lemma:frcont}.
Write $\delta^*_\mathbf{r} \equiv \lim_{k \rightarrow \infty} \delta_{k,\mathbf{r}}$, and let $\{\mathbf{x}_a: a \in \mathbb{Z}^+\}$ be a sequence in $\mathcal{P}(\mathbf{p}, \Sigma)$ with $\lim_{a \rightarrow \infty} \mathbf{x}_a = \mathbf{p}$. There are two cases. 

First suppose that $\delta^*_\mathbf{r}>0$. 
Let $\mathbf{x} \in \mathcal{P}(\mathbf{p},\delta^*_\mathbf{r}, \Sigma)$; then $\mathbf{x} \in \mathcal{P}_{k, \mathbf{r}}$ for all $k \in \mathbb{Z}^+$, so the Gittins search sequence against $\mathbf{x}$ which breaks ties using rule $\mathbf{r}$ is identical to $\xi(\mathbf{r},\mathbf{p})$. It follows that $f$ is constant on $\mathcal{P}(\mathbf{p},\delta^*_\mathbf{r}, \Sigma)$ when its first argument is fixed at $\mathbf{r}$. Furthermore, since $\lim_{a \rightarrow \infty} \mathbf{x}_a = \mathbf{p}$, there must exist $A$ such that $\mathbf{x}_a \in B(\mathbf{p}, \delta^*_\mathbf{r})$ for all $a \geq A$. Yet, since $\{\mathbf{x}_a\}$ is a sequence in $\mathcal{P}(\mathbf{p}, \Sigma)$, we also, for all $a \geq A$, have $\mathbf{x}_a \in \mathcal{P}(\mathbf{p},\delta^*_\mathbf{r}, \Sigma)$ and hence $f(\mathbf{r},\mathbf{x}_a)=f(\mathbf{r},\mathbf{p})$ for any $\mathbf{r}$ with all elements in $\Sigma$; proving the result for the first case.

Second, suppose that $\delta^*_\mathbf{r}=0$. As in the proof of Lemma \ref{lemma:frcont}, since $\{\mathbf{x}_a\}$ has limit $\mathbf{p}$ and, for any $k \in \mathbb{Z}^+$, $\delta_{k,\mathbf{r}}>0$, there must exist a smallest number $g_{\mathbf{r}}(k) \in \mathbb{Z}^+$ such that every term in $\{\mathbf{x}_a\}$ after $\mathbf{x}_{g(k)}$ belongs to the ball $B(\mathbf{p},\delta_{k,\mathbf{r}})$. Further, since $\{\mathbf{x}_a\}$ is a sequence in $\mathcal{P}(\mathbf{p}, \Sigma)$, every term in $\{\mathbf{x}_a\}$ after $\mathbf{x}_{g(k)}$ also belongs to $\mathcal{P}(\mathbf{p},\delta_{k,\mathbf{r}}, \Sigma)$. Formally, for any $k \in \mathbb{Z}^+$, we write
\begin{equation} \label{eqn:gr(k)def}
g_{\mathbf{r}}(k) \equiv \min \{A:\mathbf{x}_a \in \mathcal{P}(\mathbf{p},\delta_{k,\mathbf{r}}, \Sigma), \;a \geq A\}.
\end{equation}
Note from \eqref{eqn:delta_k,r} that since $\mathcal{P}_{k+1, \mathbf{r}} \subseteq \mathcal{P}_{k, \mathbf{r}}$, 
we have $\delta_{k, \mathbf{r}} \geq \delta_{k+1, \mathbf{r}}$. It follows that $B(\mathbf{p},\delta_{k+1, \mathbf{r}}) \subseteq B(\mathbf{p},\delta_{k,\mathbf{r}})$, and hence $g_{\mathbf{r}}(k) \leq g_{\mathbf{r}}(k+1)$ for all $k \in \mathbb{Z}^+$, so the sequence $\{g_{\mathbf{r}}(k): k \in \mathbb{Z}^+\}$ increases weakly.
An identical argument to that in the proof of Lemma \ref{lemma:frcont} for $\{\mathbf{x}_{g(k)}\}$ can be applied to $\{\mathbf{x}_{g_{\mathbf{r}}(k)}\}$ to show that $\lim_{k\rightarrow \infty} \mathbf{x}_{g_{\mathbf{r}}(k)} = \mathbf{p}$.

By the definitions in \eqref{eqn:delta_k,r} and \eqref{eqn:gr(k)def}, we have $\mathbf{x}_{g_{\mathbf{r}}(k)} \in \mathcal{P}(\mathbf{p}, \delta_{k,\mathbf{r}}, \Sigma) \subseteq \mathcal{P}_{k, \mathbf{r}}$. 
Recall $\xi(\mathbf{r},\mathbf{x})$ is the Gittins search sequence against $\mathbf{x}$ which breaks ties using rule $\mathbf{r}$. Since $g_{\mathbf{r}}$ is increasing and $\mathbf{x}_{g_{\mathbf{r}}(k)} \in \mathcal{P}_{k,\mathbf{r}}$, as $k$ increases, the first time when $\xi(\mathbf{r},\mathbf{x}_{g_{\mathbf{r}}(k)})$ and $\xi(\mathbf{r}, \mathbf{p})$ differ becomes increasingly later and later into the search. Hence, no matter where the hider is hidden, the effect on the expected time to detection of this difference decreases to 0. Therefore, $V_i(\xi(\mathbf{r},\mathbf{x}_{g_{\mathbf{r}}(k)})) \rightarrow V_i(\xi(\mathbf{r},\mathbf{p}))$ as $k \rightarrow \infty$, $i = 1,\ldots , n$. Combined with $\lim_{k\rightarrow \infty} \mathbf{x}_{g_{\mathbf{r}}(k)} = \lim_{a\rightarrow \infty} \mathbf{x}_{a} = \mathbf{p}$, we have
$$\lim_{k \rightarrow \infty} f (\mathbf{r},\mathbf{x}_{g_{\mathbf{r}}(k)}) = \lim_{a \rightarrow \infty} f(\mathbf{r}, \mathbf{x}_a) = f(\mathbf{r},\mathbf{p}),$$ for any $\mathbf{r}$ with all elements in $\Sigma$, proving the result for the second case.
\end{proof}

The continuity of $f$ in its second argument is key in proving the main result: that $\bar{\mathcal{S}}(\epsilon)$ is closed.


\paragraph{Proof that $\bar{\mathcal{S}}(\epsilon)$ is closed.}
Write $\{\mathbf{s}_a: a \in \mathbb{Z}^+\}$ for a convergent sequence in $\bar{\mathcal{S}}(\epsilon)$, and write $\mathbf{s}_0 \equiv \lim_{a \rightarrow \infty} \mathbf{s}_a$ for its limit. To show that $\bar{\mathcal{S}}(\epsilon)$ is closed, we need to show that $\mathbf{s}_0 \in \bar{\mathcal{S}}(\epsilon)$.

Since each element of $\bar{\mathcal{S}}(\epsilon)$ corresponds to some mixed hiding strategy $\mathbf{p} \in \mathcal{P}(\epsilon)$ and tie-breaking rule $\mathbf{r} \in \mathcal{R}$, $\bar{\mathcal{S}}(\epsilon)$ is equal to the image of $f$.
Therefore, for all $a \in \mathbb{Z}^+$, we may choose $\mathbf{x}_a \in \mathcal{P}(\epsilon)$ and $\mathbf{r}_a \in \mathcal{R}$ such that $\mathbf{s}_a=f(\mathbf{r}_a,\mathbf{x}_a)$. 
Further, since $\mathcal{P}(\epsilon)$ is bounded, the sequence $\{\mathbf{x}_a: a \in \mathbb{Z}^+\}$ has a convergent subsequence $\{\mathbf{x}_{h(a)}: a \in \mathbb{Z}^+\}$, and, since $\mathcal{P}(\epsilon)$ is closed, $\mathbf{x}_0 \equiv \lim_{a \rightarrow \infty} \mathbf{x}_{h(a)} \in \mathcal{P}(\epsilon)$.
Any infinite subsequence of the convergent sequence $\{\mathbf{s}_a\}$ must converge to the same limit as $\{\mathbf{s}_a\}$; therefore, we have $\mathbf{s}_0 = \lim_{a \rightarrow \infty} \mathbf{s}_{h(a)}$.



We consider two cases. First, suppose that $\{\mathbf{x}_{h(a)}\}$ attains its limit $\mathbf{x}_0$. In other words, there exists $A \in \mathbb{Z}^+$ such that $\mathbf{x}_{h(a)}=\mathbf{x}_0$ for all $a \geq A$. In this instance, the sequence $\{\mathbf{s}_{h(a)} : a \geq A\}$, which has limit $\mathbf{s}_0$, is a sequence in the set $\mathcal{S}_{\mathbf{x}_0}$ defined in \eqref{eqn:S(p)defn}, which was shown to be closed by Lemma \ref{lemma:fixedpclosed}. Therefore, $\mathbf{s}_0 \in \mathcal{S}_{\mathbf{x}_0} \subset \bar{\mathcal{S}}(\epsilon)$, completing the proof for the first case.

Second, suppose that $\{\mathbf{x}_{h(a)}\}$ does not attain its limit $\mathbf{x}_0$. Since $\mathbf{x}_0 \in \mathcal{P}(\epsilon)$, we have $f(\mathbf{r},\mathbf{x}_0) \in \bar{\mathcal{S}}(\epsilon)$ for any $\mathbf{r} \in \mathcal{R}$. To complete the proof for the second case, we show that $\mathbf{s}_0 = f(\mathbf{r},\mathbf{x}_0)$ for some $\mathbf{r} \in \mathcal{R}$.
To do this, we split the second case into two subcases.

First, consider the easier subcase in which $\mathbf{x}_0$ is a non-tie point. Then, for any $\mathbf{r} \in \mathcal{R}$, we have 
$$\mathbf{s}_0=\lim_{a \rightarrow \infty} \mathbf{s}_{h(a)} =\lim_{a \rightarrow \infty} f(\mathbf{r}_{h(a)},\mathbf{x}_{h(a)}) =  f(\mathbf{r},\mathbf{x}_0),
$$ 
where the last equality follows by Lemma \ref{lemma:frcont} and the fact that the $f(\mathbf{r},\mathbf{x}_0)$ are equal for all $\mathbf{r}\in \mathcal{R}$.

The rest of the proof concerns the more challenging subcase, in which $\mathbf{x}_0 \equiv(x_{0,1}, \ldots , x_{0,n})$ is a tie point. Note that, for any $\mathbf{x} \equiv(x_1,\ldots,x_n) \in \mathcal{P}$, there exists a subset $\Sigma_{\mathbf{x}} \subseteq S_n$ for which $\sigma \in \Sigma_{\mathbf{x}}$ if and only if $$x_{\sigma(i)}/x_{0,\sigma(i)} \geq x_{\sigma(i+1)}/x_{0,\sigma(i+1)}, \; i=1,2,\ldots, n-1. $$
Recalling the definition in \eqref{eqn:Psigmaset}, we have $\mathbf{x} \in \mathcal{P}(\mathbf{x}_0,\Sigma_\mathbf{x})$, and further $\Sigma_\mathbf{x}$ is the unique subset of maximal size such that $\mathbf{x} \in \mathcal{P}(\mathbf{x}_0,\Sigma_\mathbf{x})$. 
Since the elements $\{x_{i}/x_{0,i}, \; i = 1, \ldots, n\}$ must lie in some size order, $\Sigma_\mathbf{x}$ is non-empty for any $\mathbf{x} \in \mathcal{P}(\epsilon)$.

Since there are a finite number of subsets of $S_n$, there must exist $\Sigma^* \subset S_n$ and a convergent subsequence of $\{\mathbf{x}_{h(a)}\}$, say $\{\mathbf{x}_m : m \in \mathbb{Z}^+\}$, such that $\Sigma_{\mathbf{x}_m} = \Sigma^*$ for all $m \in \mathbb{Z}^+$. In other words, $\{\mathbf{x}_m\}$ is a sequence in $\mathcal{P}(\mathbf{x}_0, \Sigma^*)$.

Since $\{\mathbf{x}_m\}$ is a convergent subsequence of $\{\mathbf{x}_{h(a)}\}$, we have $\lim_{m \rightarrow \infty} \mathbf{x}_{m} = \mathbf{x}_0$, and
\begin{equation*} 
\lim_{m \rightarrow \infty} f(\mathbf{r}_{m},\mathbf{x}_{m}) = \lim_{m \rightarrow \infty} \mathbf{s}_{m} = \mathbf{s}_0.
\end{equation*}

To finish the proof, which, recall, involves showing $\mathbf{s}_0 = f(\mathbf{r},\mathbf{x}_0)$ for some $\mathbf{r} \in \mathcal{R}$, we split into two further subcases, numbered below.
\begin{enumerate}
\item First, we consider the easier subcase in which there are finitely many tie points in $\{\mathbf{x}_m\}$. 
In this case, we may choose $M$ such that there are no tie points in the sequence $\{\mathbf{x}_{m}: m \geq M\}$. Therefore, for all $m \geq M$, we have $f(\mathbf{r}_{m},\mathbf{x}_{m}) = f(\mathbf{r},\mathbf{x}_{m})$ for all $\mathbf{r} \in \mathcal{R}$. Let $\mathbf{r}^* \in \mathcal{R}$ contain only elements in $\Sigma^*$. Then we have
$$\mathbf{s}_0 = \lim_{m \rightarrow \infty} f(\mathbf{r}_{m},\mathbf{x}_{m}) = \lim_{m \rightarrow \infty} f(\mathbf{r}^*,\mathbf{x}_{m}) = f(\mathbf{r}^*,\mathbf{x}_0),$$
where the last equality follows by Lemma \ref{lem:frsemicont}.

\item Now suppose that there are infinitely many tie points in $\{\mathbf{x}_m\}$. We begin with two observations for any $\mathbf{x} \in \mathcal{P}(\epsilon)$.

\paragraph{Observation 1:} First, note that the position of any equalities in the ordering of the terms $\{x_{i}/x_{0,i}, \; i = 1, \ldots, n\}$ completely determines $\Sigma_{\mathbf{x}}$. In particular, for any pair of boxes $i,j \in \{1,\ldots , n\}$, we have $x_i/x_{0,i} =x_j/x_{0,j}$ if and only if there exists $\sigma_1,\sigma_2 \in \Sigma_\mathbf{x}$ with $\sigma_1(i)>\sigma_1(j)$ and  $\sigma_2(j)>\sigma_2(i)$. Also, we have $x_i/x_{0,i} > x_j/x_{0,j}$ if and only if $\sigma(i)>\sigma(j)$ for all $\sigma \in \Sigma_\mathbf{x}$.

\paragraph{Observation 2:} Second, for any pair of boxes $i,j \in \{1,\ldots , n\}$ and any $y,z \in \mathbb{N}$, where $\mathbb{N}$ is the set of nonnegative integers, write
$$k_{i,j}(y,z) \equiv \frac{q_j(1-q_j)^{z}t_i}{q_i(1-q_i)^{y}t_j},$$
recalling that $q_i$ (resp. $t_i$) is the detection probability (resp. search time) of box $i$, $i=1,\ldots , n$.
Then, inspection of \eqref{eq:simpleGI} shows that, following a Gittins search sequence against $\mathbf{x}$, there is a tie between boxes $i$ and $j$ after $y$ (resp. $z$) searches of box $i$ (resp. $j$) have been made if and only if $x_i/x_j = k_{i,j}(y,z)$.

\paragraph{End of the Proof:}
Now consider the sequence $\{\mathbf{x}_m\}$, and write $\mathbf{x}_m\equiv(x_{m,1},\ldots, x_{m,n})$ for the $m$th term, $m \in \mathbb{Z}^+$. Since $\{\mathbf{x}_m\}$ has limit $\mathbf{x}_0$,  
for any two boxes $i,j \in \{1,\ldots, n\}$, we have
\begin{equation} \label{eqn:xmlimit}
\frac{x_{m,i}}{x_{m,j}} \rightarrow \frac{x_{0,i}}{x_{0,j}} \quad \text{as} \quad m \rightarrow \infty.
\end{equation}

Now choose $\mathbf{x} \in \{\mathbf{x}_m\}$ and suppose that $c,d \in \{1,\ldots,n\}$ satisfy $x_c/x_{0,c} \neq x_d/x_{0,d}$. Then, by Observation 1, the same must be true for every element of $\{\mathbf{x}_m\}$ since $\Sigma_{\mathbf{x}_m}=\Sigma^*$ for all $m \in \mathbb{Z}^+$. Hence, the limit in \eqref{eqn:xmlimit} is never attained for $i=c$ and $j=d$. In other words, $x_{m,c}/x_{m,d}$ approaches but never reaches $x_{0,c}/x_{0,d}$ as $m \rightarrow \infty$.

Let $y, z \in \mathbb{N}$ and consider $k_{c,d}(y,z)$ defined in Observation 2. There are two scenarios. First, we may have $k_{c,d}(y,z)=x_{0,c}/x_{0,d}$; in this scenario, since the limit in \eqref{eqn:xmlimit} is never attained, in no Gittins search sequence against any element of $\{\mathbf{x}_m\}$ is there a tie between boxes $c$ and $d$ after $y$ (resp. $z$) searches of box $c$ (resp. $d$) have been made. Second, if $k_{c,d}(y,z) \neq x_{0,c}/x_{0,d}$, by the limit in \eqref{eqn:xmlimit}, there exists a finite smallest element of $\mathbb{Z}^+$, say $M_{c,d}(y,z)$, such that the same statement holds after the $M_{c,d}(y,z)$th term of $\{\mathbf{x}_m\}$; in other words, in no Gittins search sequence against any element of $\{\mathbf{x}_m: M_{c,d}(y,z) \geq m\}$ is there a tie between boxes $c$ and $d$ after $y$ (resp. $z$) searches of box $c$ (resp. $d$) have been made. 

For any $b \in \mathbb{Z}^+$, write
$$M^b_{c,d} \equiv \max \{M_{c,d}(y,z) : y, z \in \mathbb{N} \; \text{with} \; y+z \leq b \}.$$ 
It follows that, whilst the total number of searches of boxes $c$ and $d$ is no larger than $b$, no ties involving \emph{both} boxes $c$ and $d$ are encountered in a Gittins search sequence against any element of $\{\mathbf{x}_m: m \geq M^b_{c,d}\}$. 
Clearly $\{M^b_{c,d} : b \in \mathbb{Z}^+\}$ forms an increasing sequence; therefore, as $m \rightarrow \infty$, any tie involving both boxes $c$ and $d$ occurs increasingly later and later into the search, so the effect on the expected time to detection of how such a tie is broken decreases to 0.

Therefore, as $m \rightarrow \infty$, only the manner in which ties involving \emph{only} boxes $i$ and $j$ satisfying $x_i/x_{0,i} = x_j/x_{0,j}$ 
are broken has any effect on the expected time to detection under a Gittins search sequence against an element of $\{\mathbf{x}_m\}$.
Without a loss of generality, suppose such a tie is between boxes $1,\ldots , y$, for some $y \in \{2,\ldots, n\}$. Suppose the tie is broken by $\sigma \in S_n$. By Observation 1, since $x_1/x_{0,1} = \cdots = x_y/x_{0,y}$, there exists $\sigma^* \in \Sigma_\mathbf{x}$ which ranks boxes $1,\ldots , y$ in the same order as $\sigma$. Therefore, breaking the tie using $\sigma^*$ leads to boxes $1,\ldots , y$ being searched in the same order as breaking the tie using $\sigma$. It follows that there exists $\mathbf{r}^*$ with only elements in $\Sigma^*$ such that, if, for all $m \in \mathbb{Z}^+$, we replace $\mathbf{r}_m$ with $\mathbf{r}^*$, as $m \rightarrow \infty$, the effect on the expected time to detection tends to 0. In other words,
$$\mathbf{s}_0 = \lim_{m \rightarrow \infty} f(\mathbf{r}_{m},\mathbf{x}_{m}) = \lim_{m \rightarrow \infty} f(\mathbf{r}^*,\mathbf{x}_{m}) = f(\mathbf{r}^*,\mathbf{x}_0),$$
where the last equality follows by Lemma \ref{lem:frsemicont}. The proof is completed. \ \rule{0.5em}{0.5em}
\end{enumerate}


\section{Gittins' Proposition 8.5 Extension} \label{append:Git}
For any (pure or mixed) search strategy $\eta$ and $i,j \in \{1,\ldots , n\}$, write $D_{i,j}(\eta) \equiv V_i(\eta)-V_j(\eta)$.
If $D_{i,j}(\eta)=0$ for some $i,j \in \{1,\ldots , n\}$, we say $\eta$ \emph{equalizes} boxes $i$ and $j$. Let $\eta_1$ and $\eta_2$ be search strategies, and choose $i,j \in \{1,\ldots,n\}$ such that $D_{i,j}(\eta_1)\leq 0$. Then, by direct computation, it is easy to show that there exists a mixture of $\eta_1$ and $\eta_2$ that equalizes boxes $i$ and $j$ if any only if $D_{i,j}(\eta_2)\geq 0$.

Let $\mathbf{p}^*$ be an arbitrary optimal hiding strategy.
Recall $\xi_{\sigma,\mathbf{p}^*}$ is the Gittins search sequence against $\mathbf{p}^*$ that breaks every tie encountered using $\sigma$; write $\xi_{\sigma} \equiv \xi_{\sigma,\mathbf{p}^*}$.
When $n=2$, we have $\widehat{\mathcal{C}}_{\mathbf{p}^*}=\{\xi_{12}, \xi_{21}\}$, where $\xi_{12}$ (resp. $\xi_{21}$) breaks any tie in favour of box 1 (resp. box 2). The proof of Gittins' Lemma 8.4 shows that $D_{1,2}(\xi_{12}) \leq 0$ and $D_{1,2}(\xi_{21}) \geq 0$. In other words, there exists a mixture $\eta^*$ of $\xi_{12}$ and $\xi_{21}$ 
which equalizes boxes 1 and 2; by Theorem \ref{thm:Git8.3}, $\eta^*$ is optimal for the searcher.

By extending Gittins' method to an $n$-box problem, for any $i,j \in \{1,\ldots,n\}$, we can show that  $D_{i,j}(\xi_{i\cdots j}) \leq 0$ and $D_{i,j}(\xi_{j\cdots i}) \geq 0$, where $x\cdots y$ is any permutation of $\{1,\ldots , n\}$ with first element $x$ and last element $y$. If either of these two $D_{i,j}$ terms is equal to 0, then the corresponding search sequence equalizes $i$ and $j$. Otherwise, since, for any $\xi \in \mathcal{C}_{\mathbf{p}^*}$, $D_{i,j}(\xi)$ must lie one side of 0, there exists a mixture of $\xi$ and either $\xi_{i\cdots j}$ or $\xi_{j\cdots i}$ that equalizes boxes $i$ and $j$. Therefore, many mixtures of pairs of sequences in $\mathcal{C}_{\mathbf{p}^*}$ can be constructed that equalize boxes $i$ and $j$.


To obtain an optimal search strategy using Theorem \ref{thm:Git8.3}, we need a mixture of elements of $\mathcal{C}_{\mathbf{p}^*}$ that equalizes all $n$ boxes. Yet, problems occur when a third box, say $k$, is introduced. 
Suppose $\eta_1$ and $\eta_2$ both equalize boxes $i$ and $j$; then, any mixture of $\eta_1$ and $\eta_2$ that equalizes boxes $i$ and $k$ (or $j$ and $k$) will equalize boxes $i$, $j$ and $k$. However, there is no guarantee that $D_{i,k}(\eta_1)$ and $D_{i,k}(\eta_2)$ will have opposing signs, so there is no guarantee that such a mixture exists. 
Whilst we managed to prove that such $\eta_1$ and $\eta_2$ 
with $D_{i,k}(\eta_1)\leq 0 \leq D_{i,k}(\eta_2)$ exist when $n=3$ (thus finding an optimal search strategy for the three-box case), the proof cannot be generalized to $n \geq 4$.

\section{Details of Expected Time to Detection Calculations} \label{append:numerdetails}
In this appendix section, we provide extra details of the methods described in Section \ref{sec:numercalcs} to calculate expected times to detection under a Gittins search sequence in cyclic and acyclic games.

\paragraph{A Closed Form for an Cyclic Game}
Consider a cyclic search game, which satisfies \eqref{eqn:cyclic}. By \cite{Matula}, after an initial transient period, any Gittins search sequence $\xi$ will make $\widehat{x} \equiv \sum_{i=1}^n x_i$ consecutive searches involving exactly $x_i$ visits of box $i$ for $i=1,\ldots , n$. 
By \eqref{eqn:cyclic}, the posterior probabilities that the hider is in each box will be the same before and after these $\widehat{x}$ searches have been made. Therefore, $\xi$ will cycle these $\widehat{x}$ searches indefinitely; 
such repetition allows a closed form for $V_i(\xi)$, $i=1,\ldots , n$, to be calculated as follows.

Suppose $\xi$ has entered the cycle of $\widehat{x}$ searches after some $K$ searches of box $i$ have been made. Therefore, for any $a, k \in \{0,1,2,\ldots\}$, we have $b_i(K+k+ax_i, \xi) = b_i(K+k, \xi) + a\widehat{x}$, where, recall, $b_i(z,\xi)$ is the time at which the $z$th search of box $i$ is made under $\xi$. Write $V_K \equiv \sum_{k=1}^{K} (1-q_i)^{k-1} [b_i(k,\xi)-b_i(k-1,\xi)]$ for the finite approximation to \eqref{eqn:genv_i(xi)} after $K$ searches of box $i$. We have
\begin{equation*}
V_i(\xi) = V_K + (1-q_i)^K \sum_{a=0}^\infty \left( \sum_{k=1}^{x_i} (1-q_i)^{ax_i+k-1} [b_i(K+k+ax_i,\xi)-b_i(K+k+ax_i-1,\xi)] \right),
\end{equation*}
from which it follows that

\begin{align*}
\frac{V_i(\xi) -V_K}{(1-q_i)^K} 
&= \sum_{a=0}^\infty (1-q_i)^{ax_i} \left( \sum_{k=1}^{x_i} (1-q_i)^{k-1} \left[b_i(K+k, \xi) - b_i(K+k-1, \xi)\right] \right) \\
&=\frac{A_i(K,\xi)}{(1-(1-q_i)^{x_i})},
\end{align*}
where 
\begin{equation*} 
A_i(K,\xi) \equiv \sum_{k=1}^{x_i} (1-q_i)^{k-1} \left[b_i(K+k, \xi) - b_i(K+k-1, \xi)\right].
\end{equation*}
Therefore, to evaluate $V_i(\xi)$ precisely, we only need to calculate $b_i(k,\xi)$ for those $k \in \{1,\ldots, K+x_i\}$, which we discuss in the following.

As justified in Section \ref{sec:numercalcs}, for $\mathbf{p}$ a solution to a finite matrix game in Algorithm \ref{al:minExpCost}, using floating-point numbers to compute the Gittins indices in \eqref{eq:simpleGI} will calculate $b_i(k,\xi)$ for $k \in \{1,2,\ldots\}$ for some $\xi \in \mathcal{C}_{\mathbf{p}}$, which is sufficient for step \ref{step:check_stop} of Algorithm \ref{al:minExpCost} and \ref{stepei:update} of Algorithm \ref{al:get_interior}. To determine $K$, we 
evaluate $\xi$ 
until $\widehat{x}$ consecutive searches involve $x_j$ searches of box $j$ for $j=1,\ldots, n$.

On the other hand, floating-point indices cannot reliably calculate specific search sequences in $\mathcal{C}_{\mathbf{p}_0}$, a requirement of step \ref{step:setup} of Algorithm \ref{al:minExpCost}. Therefore, to calculate $b_i(k,\xi)$ for a specific $\xi \in \mathcal{C}_{\mathbf{p}_0}$, a set of alternative indices is derived below, which encodes integers rather than floating-point numbers.


First, note that, by \eqref{eqn:cyclic}, the first $\widehat{x}$ searches of any $\xi \in \mathcal{C}_{\mathbf{p}_0}$ will involve $x_j$ searches of box $j$, $j=1,\ldots , n$. Therefore, we may take $K=0$, so only need calculate $b_i(k,\xi)$ for $k \in \{1,\ldots, x_i\}$.
For any $\xi \in \mathcal{C}_{\mathbf{p}_0}$, all $n$ indices in \eqref{eq:simpleGI} are equal at the start of the search, say to $y$. 
For $i=1,\ldots , n$, suppose $m_i \in \{1,\ldots, x_i\}$ searches of box $i$ have been made, so the current corresponding index in \eqref{eq:simpleGI} is $y(1-q_i)^{m_i}$. Then we have $$y(1-q_i)^{m_i} \propto (1-q_i)^{m_i} = \left((1-q_i)^{x_i}\right)^{m_i/x_i}= c^{m_i/x_i} \propto x_i/m_i.$$
Therefore, the rule in \eqref{eq:simpleGI} is equivalent to searching any box $j$ satisfying
\begin{equation} \label{eq:cyclicGI}
j = \argmax_{i \in \{1, \ldots ,  n\}} \frac{x_i}{m_i}.
\end{equation}
Yet, both $x_i$ and $m_i$ are integers, so, unlike using \eqref{eq:simpleGI}, ties will always be detected using \eqref{eq:cyclicGI}.

\paragraph{An Upper Bound for an Acyclic Game}
For a Gittins search sequence $\xi$, whilst a finite approximation to \eqref{eqn:genv_i(xi)} after any $K\in\{1,2,\ldots\}$ searches of box $i$ gives a lower bound for $V_i(\xi)$, we calculate an upper bound for $V_i(\xi)$ using the following method.

Write
\begin{equation*}
m \equiv \floor[\Bigg]{\max_{i,j \in \{1,\ldots , n\}} \frac{\log(1-q_i)}{\log(1-q_j)}} + 1.
\end{equation*}
For any $i,j \in \{1,\ldots, n\}$, since $m > \log(1-q_i)/\log(1-q_j)$, then $(1-q_j)^m < (1-q_i)$, so between any two successive searches of box $i$, any Gittins search sequence will make at most $m$ searches of box $j$, for $i \neq j$.
Therefore, for any $i \in \{1,\ldots,n\}$, no more than time $\widehat{m} \equiv \sum_{j=1}^n mt_j$ can elapse between successive searches of box $i$ following any Gittins search sequence. 


It follows that, for any Gittins search sequence $\xi$ and $K \in \{1,2,\ldots\}$, following $\xi$ until $K$ searches of box $i$ have been made, then, after that, assuming box $i$ is searched at regular time intervals of length $\widehat{m}$ gives an upper bound on $V_i(\xi)$. In other words,
\begin{align} \label{eqn:upperboundV}
V_i(\xi) &\leq \sum_{k=1}^{K} (1-q_i)^{k-1} [b_i(k,\xi)-b_i(k-1,\xi)] + (1-q_i)^{K} \sum_{k=1}^{\infty} \widehat{m}(1-q_i)^{k-1}  \nonumber \\
&= \sum_{k=1}^{K} (1-q_i)^{k-1} [b_i(k,\xi)-b_i(k-1,\xi)] + \frac{\widehat{m}(1-q_i)^{K}}{q_i}.
\end{align} 
Note that the first term in \eqref{eqn:upperboundV} is the lower bound for $V_i(\xi)$ obtained via a finite approximation to \eqref{eqn:genv_i(xi)} after $K$ searches of box $i$. Therefore, we increase $K$ until the ratio of the second term of \eqref{eqn:upperboundV} divided by the first term of \eqref{eqn:upperboundV} is less than $10^{-10}$.

\end{appendices}
\bibliographystyle{apalike}
\bibliography{References}

\begin{thebibliography}{}

\bibitem[Alpern et~al., 2013]{Alpern4auth}
Alpern, S., Fokkink, R., Gasieniec, L., Lindelauf, R., and Subrahmanian, V.
  (2013).
\newblock {\em Search Theory: A Game Theoretic Perspective}.
\newblock Springer Publishing Company, Incorporated.

\bibitem[Alpern and Gal, 2003]{AlpernGal}
Alpern, S. and Gal, S. (2003).
\newblock {\em The Theory of Search Games and Rendezvous}.
\newblock International Series in Operations Research and Managment Science.
  Kluwer Academic Publishers, Boston, Dordrecht, London.

\bibitem[Black, 1965]{Black:1965ts}
Black, W.~L. (1965).
\newblock {Discrete sequential search}.
\newblock {\em Information and Control}, 8(2):159--162.

\bibitem[Blackwell and Girshick, 1954]{Blackwell1954}
Blackwell, D. and Girshick, M.~A. (1954).
\newblock {\em Theory of Games and Statistical Decisions}.
\newblock John Wiley \& Sons, New York.

\bibitem[Bram, 1963]{Bram1963}
Bram, J. (1963).
\newblock A 2-player n-region search game.
\newblock IRM-31, Operations Evaluation Group, Center for Naval Analysis.

\bibitem[Charalambos and Aliprantis, 2013]{Chara:2013}
Charalambos, D. and Aliprantis, B. (2013).
\newblock {\em Infinite Dimensional Analysis: A Hitchhiker's Guide}.
\newblock Springer-Verlag Berlin and Heidelberg GmbH \& Company KG.

\bibitem[Chew, 1973]{ChewJr:1973td}
Chew, Jr, M.~C. (1973).
\newblock {Optimal stopping in a discrete search problem}.
\newblock {\em Operations Research}, 21(3):741--747.

\bibitem[Clarkson et~al., 2020]{Clarkson:2020}
Clarkson, J., Glazebrook, K.~D., and Lin, K.~Y. (2020).
\newblock Fast or slow: Search in discrete locations with two search modes.
\newblock {\em Operations Research}, 68(2):552--571.

\bibitem[Ferguson, 2020]{Ferguson2020}
Ferguson, T.~S. (2020).
\newblock {\em A Course in Game Theory}.
\newblock WSPC.

\bibitem[Garrec and Scarsini, 2020]{Garrec:2020}
Garrec, T. and Scarsini, M. (2020).
\newblock Search for an immobile hider on a stochastic network.
\newblock {\em European Journal of Operational Research}, 283(2):783--794.

\bibitem[Gittins and Roberts, 1979]{GitRob2}
Gittins, J. and Roberts, D. (1979).
\newblock The search for an intelligent evader concealed in one of an arbitrary
  number of regions.
\newblock {\em Naval Research Logistics Quarterly}, 26(4):651--666.

\bibitem[Gittins et~al., 2011]{KGbook}
Gittins, J., Weber, R., and Glazebrook, K.~D. (2011).
\newblock {\em Multi-Armed Bandit Allocation Indices}.
\newblock John Wiley and Sons, Ltd.

\bibitem[Gittins, 1979]{Gittins79}
Gittins, J.~C. (1979).
\newblock Bandit processes and dynamic allocation indices.
\newblock {\em Journal of the Royal Statistical Society. Series B
  (Methodological)}, 41(2):148--177.

\bibitem[Gittins, 1989]{Gittins:1989}
Gittins, J.~C. (1989).
\newblock {\em Multi-armed bandit allocation indices}.
\newblock Wiley.

\bibitem[Kadane, 1971]{Kadane:1971wj}
Kadane, J.~B. (1971).
\newblock {Optimal whereabouts search}.
\newblock {\em Operations Research}, 19(4):894--904.

\bibitem[Kress et~al., 2008]{Kress:2008co}
Kress, M., Lin, K.~Y., and Szechtman, R. (2008).
\newblock {Optimal discrete search with imperfect specificity}.
\newblock {\em Mathematical Methods Of Operations Research}, 68(3):539--549.

\bibitem[Lin and Singham, 2016]{Lin2016}
Lin, K.~Y. and Singham, D.~I. (2016).
\newblock Finding a hider by an unknown deadline.
\newblock {\em Oper. Res. Lett.}, 44(1):25--32.

\bibitem[Matula, 1964]{Matula}
Matula, D. (1964).
\newblock A periodic optimal search.
\newblock {\em The American Mathematical Monthly}, 71(1):15--21.

\bibitem[Norris, 1962]{Norris1962}
Norris, R.~C. (1962).
\newblock Studies in search for a conscious evader.
\newblock {\em MIT Lincoln Laboratory Technical Report Number 279}.

\bibitem[Roberts and Gittins, 1978]{GitRob1}
Roberts, D.~M. and Gittins, J.~C. (1978).
\newblock The search for an intelligent evader: Strategies for searcher and
  evader in the two-region problem.
\newblock {\em Naval Research Logistics Quarterly}, 25(1):95--106.

\bibitem[Ross, 1969]{Ross:1969cm}
Ross, S.~M. (1969).
\newblock {A Problem in Optimal Search and Stop}.
\newblock {\em Operations Research}, 17(6):984--992.

\bibitem[Ross, 1983]{Ross:1983}
Ross, S.~M. (1983).
\newblock {\em Introduction to Stochastic Dynamic Programming: Probability and
  Mathematical}.
\newblock Academic Press, Inc., USA.

\bibitem[Ruckle, 1991]{Ruckle:1991}
Ruckle, W.~H. (1991).
\newblock A discrete search game.
\newblock In {\em Stochastic Games And Related Topics}, pages 29--43. Springer.

\bibitem[Stone, 2004]{Stone2004}
Stone, L.~D. (2004).
\newblock {\em Theory of Optimal Search}.
\newblock INFORMS, 2nd edition.

\bibitem[Stone et~al., 2016]{Stone2016}
Stone, L.~D., Royset, J.~O., and Washburn, A.~R. (2016).
\newblock {\em Optimal Search for Moving Targets}.
\newblock International Series in Operations Research and Management Science.
  Springer, New York, United States.

\bibitem[Subelman, 1981]{Subelman}
Subelman, E.~J. (1981).
\newblock A hide-search game.
\newblock {\em Journal of Applied Probability}, 18:628--640.

\bibitem[Washburn, 2002]{Washburn:2002tw}
Washburn, A.~R. (2002).
\newblock {\em Search and Detection}.
\newblock INFORMS, 4th edition.

\bibitem[Washburn, 2003]{Washburn:2003un}
Washburn, A.~R. (2003).
\newblock {\em {Two-Person Zero-Sum Games}}.
\newblock INFORMS, Rockville, MD, 3rd edition.

\bibitem[Wegener, 1980]{Wegener:1980hv}
Wegener, I. (1980).
\newblock {The Discrete Sequential Search Problem with Nonrandom Cost and
  Overlook Probabilities}.
\newblock {\em Mathematics Of Operations Research}, 5(3):373--380.

\end{thebibliography}

\end{document}